## Neural-Bayesian Structure Learning for Discrete Choice Modeling

**Hyunsoo Yun[1†], Eun Hak Lee[2†], Jiaru Zhang[1], Ziran Wang[1], and Eui-Jin Kim[3*]**

[1]College of Engineering, Purdue University, United States of America

[2]School of Railway Operation Systems, Kyungil University, Republic of Korea

[3]Department of Transportation Systems Engineering, Ajou University, Republic of Korea

[*]Corresponding author (euijin@ajou.ac.kr)

[†]These authors contributed equally to this work.

---

### CRediT authorship contribution statement

**Hyunsoo Yun**: Writing – original draft, Writing – review & editing, Conceptualization, Methodology, Data curation, Software, Formal analysis, Visualization. **Eun Hak Lee**: Writing – review & editing, Conceptualization, Formal analysis, **Jiaru Zhang:** Writing – review & editing**,** Formal analysis, **Ziran Wang:** Writing – review & editing, Formal analysis**, Eui-Jin Kim:** Writing – review & editing, Conceptualization, Methodology, Data curation, Formal analysis, Supervision

### Declaration of competing interest

The authors declare that they have no known competing financial interests or personal relationships that could have appeared to influence the work reported in this paper.

### Acknowledgements

This work was supported by the National Research Foundation of Korea (NRF) grant funded by the Korea government (MSIT) (No.RS-2025-00520515).

### Data availability

The LPMC dataset is publicly available and documented in Hillel et al. (2018). The SP data collected in the Seoul metropolitan area is not publicly available, as the authors are not authorized to redistribute them. The source code for Neural-BSL is publicly available at https://github.com/HyunsooYun/neural-bsl.

**Abstract**

Travel mode choice is central to transportation planning, both for explaining observed choices and for anticipating mode-share responses to changes in traveler or transport-system conditions. Yet conventional discrete choice and machine learning models are estimated primarily from observational data and typically treat explanatory covariates as parallel inputs, providing no internal mechanism for determining how related attributes should adjust when one is deliberately changed. This paper proposes Neural-Bayesian Structure Learning (Neural-BSL), a framework coupling differentiable structure learning with random-utility-based discrete choice estimation in a single differentiable procedure. To prevent mutually exclusive choice outcome from distorting the recovered attribute structure, the observed choice is maintained outside the graph as an alternative-specific utility comparison, while the attribute structure and random-utility parameters are learned jointly. The learned structure enters the choice model through structure-weighted attribute interactions and provides the structural basis for propagating interventions through downstream attributes. An intervention is evaluated by updating the intervened attribute, propagating its model-implied downstream changes in topological order, and then recomputing utilities and choice probabilities. This yields both predicted mode-share responses and the associated changes in downstream traveler or trip attributes. We evaluate Neural-BSL using stated-preference data from Seoul, South Korea, and the revealed-preference London Passenger Mode Choice dataset. Neural-BSL achieves predictive performance comparable to multinomial logit and machine learning benchmarks while recovering behaviorally coherent dependency structures. Across policy scenarios, propagating interventions through the learned structure changes the predicted redistribution across modes while exposing the downstream traveler and trip adjustments underlying those responses.

## 1. Introduction

Travel mode choice is a fundamental component of travel behavior, reflecting how travelers trade time, cost, comfort, and contextual factors when selecting among available alternatives. Understanding and modeling this choice process has long been central to transportation research, as accurate representations of mode choice support both behavioral theory and a wide range of planning applications, from fare and infrastructure decisions (Guajardo Ortega and Link, 2025; Yun et al., 2024) to the modal-split stage of regional travel demand forecasting (Kim et al., 2024).

Mode choice analysis is typically conducted by estimating a model from observational data describing travelers, alternatives, and observed choices, and then using the fitted model to anticipate how mode shares would change under a proposed policy (Cascetta and Papola, 2001; El Zarwi et al., 2017). The implicit assumption underlying this workflow is that the model captures not merely what predicts mode choice, but relationships that remain valid when an explanatory variable is deliberately changed. Standard estimation procedures, however, do not guarantee this. Conventional choice specifications generally treat explanatory covariates as parallel inputs without explicitly modeling directed dependencies among traveler attributes. Models are therefore fitted to the observed joint distribution, leaving unresolved how related attributes should adjust when one of them is deliberately changed. The resulting parameters primarily describe relationships under that observed distribution rather than how choice responds to deliberate intervention.

Shmueli (2010) articulated this distinction between prediction and explanation as different statistical tasks. A predictive model may exploit any statistical regularity that improves out-of-sample accuracy under the observed distribution, whereas an explanatory model seeks relationships that provide insight into the process generating the outcome rather than predictive performance alone. Within the transportation literature specifically, Chauhan et al. (2025) revisited this distinction and observed that the boundary between predictive and causal modeling has often been blurred. Coefficients from association-based models may consequently be given causal interpretations even when the estimation procedure itself does not identify the response to intervention.

The two dominant modeling traditions in travel mode choice analysis are primarily estimated from associations observed in the data rather than from explicit causal identification. Discrete choice models grounded in random utility theory provide interpretable parameter estimates whose signs and magnitudes are routinely used in behavioral and policy analysis (Cascetta, 2009). The multinomial logit (MNL) and its mixed-logit extensions estimate utility coefficients by maximum likelihood on observational data. The estimated coefficients are interpretable in the sense that their signs and magnitudes have clear behavioral meaning under the model's assumptions, and they are often given causal interpretations in policy analysis (Kim et al., 2021; Martín-Baos et al., 2023). Machine learning approaches developed more recently, such as neural networks and tree-based ensemble methods, relax functional-form restrictions and have improved predictive accuracy on benchmark mode choice datasets (Kim and Bansal, 2024; Lee et al., 2018). Their interpretive devices, usually Shapley Additive Explanations (SHAP) values and related importance measures (Kim, 2021; Lee, 2022; Lee et al., 2022), quantify how strongly each input variable contributes to the fitted prediction.

Neither family of methods, however, identifies an interventional response from predictive fit alone. The maximum-likelihood coefficients of an MNL describe how utility varies with explanatory variables conditional on the dependence structure represented in the estimation data and the assumptions of the specification. Similarly, SHAP attributes a fitted prediction across input variables under the data-generating distribution; it does not specify how the remaining inputs should change when one of them is deliberately modified (Chauhan et al., 2025). The issue is therefore not whether these quantities are behaviorally or predictively useful, but whether the fitted model also represents how related variables adjust under deliberate intervention.

This matters for policy use because policy questions are interventional. A planner asking whether a fare reduction will increase transit ridership is not asking which feature best predicts current transit use; they are asking how ridership would respond if the fare were actually changed. The two questions can have different answers from the same data. A variable strongly associated with mode choice in observed data may have little effect under intervention if the association is driven by an unmeasured common cause, while an intervention on an upstream attribute may also change other attributes downstream of it. Consider income and vehicle ownership. If income influences vehicle ownership and both variables enter the utility specification, a simulation that changes income while holding vehicle ownership fixed evaluates the choice response conditional on observed ownership. It does not represent the additional adjustment that may occur through a change in vehicle ownership. More generally, a model whose explanatory variables enter as parallel inputs provides no internal representation of such upstream and downstream relationships. The limitation of black-box specifications such as multilayer perceptrons (MLPs) is similar: they can reproduce complex associations among inputs but provide no explicit representation of their directed relationships and no interventional operation beyond replacing values in the input vector (Natterer et al., 2025).

The natural response is to estimate causal quantities directly. This, however, is more difficult than predictive modeling. Estimating the causal effect of a variable requires comparing the outcomes that would arise under alternative values of that variable for the same observational unit; only one of these outcomes is observed in any given record, while the others must be inferred (Kim et al., 2026a). A central object for representing the assumptions underlying such inference is the directed acyclic graph (DAG), whose nodes are variables and whose directed edges encode which variables may causally influence which others. Recovering this graph from observational data, a task known as data-driven structure learning (or causal discovery), is computationally difficult because the space of candidate DAGs grows super-exponentially with the number of variables. Established structure learning algorithms, including the Peter-Clark algorithm (Spirtes and Glymour, 1991) and the GES family (Chickering, 2003), navigate this space through combinatorial search procedures guided by conditional-independence tests or likelihood-based scores.

A more recent line of work in the machine learning community has reformulated the problem as continuous optimization, in which the acyclicity constraint on the discovered graph is encoded through a smooth function differentiable in the entries of the weighted adjacency matrix. NOTEARS (Zheng et al., 2020, 2018) introduced this reformulation, often called differentiable structure learning, and DiBS (Lorch et al., 2021) subsequently extended it to a Bayesian setting in which uncertainty over graph structures is approximated through particle-based variational inference. The shared idea is to convert structure learning from a discrete search problem into gradient-based optimization, avoiding explicit enumeration of candidate graphs and making the structural parameters compatible with other differentiable model components.

Applying differentiable structure learning to mode choice modeling nevertheless presents three methodological challenges. First, random utility theory posits that each alternative carries a latent scalar utility and that the observed choice is generated by comparing these utilities (Ben-Akiva et al., 2002; Vij and Walker, 2016). Standard observed-variable structure learning formulations, by contrast, learn relationships among the variables represented as graph nodes (Spirtes et al., 2000). Treating the observed categorical choice as an ordinary structural node in place of the latent alternative utilities would therefore discard the alternative-specific random utility structure that gives the choice its behavioral meaning.

Second, the observed choice is a mutually exclusive categorical outcome: one alternative is selected from the available choice set. When represented computationally through alternative indicators, these indicators are subject to a strict deterministic constraint (i.e., their sum equals one). Introducing them as separate structural nodes can therefore generate dependencies that primarily reflect the mathematical encoding of the categorical outcome rather than substantive behavioral relationships. More generally, although structure learning methods can accommodate discrete and nonlinear variables in various ways (Andrews et al., 2018; Huang et al., 2018), the role of the observed choice differs fundamentally from that of traveler attributes because the alternatives are competing outcomes of a single choice process rather than variables whose directed relationships are being learned.

Third, evaluating the choice outcome within the structural likelihood objective can distort the dependency structure learned over explanatory attributes. Traveler attributes and the discrete choice are governed by fundamentally different data-generating processes: the former reflect socioeconomic dependencies, whereas the latter arises from utility maximization. Forcing a single structural objective to model both can therefore cause the recovered attribute DAG to reflect how the categorical choice outcome is represented, rather than the structural dependencies among traveler characteristics. The three challenges share a common origin: the choice variable is structurally different from the traveler attributes. A framework combining structure learning and discrete choice modeling should therefore retain the observed choice as the outcome of an alternative-specific utility comparison while learning the dependency structure over the explanatory attributes.

We propose **Neural-Bayesian Structure Learning (Neural-BSL)**, a framework that integrates data-driven structure learning over traveler attributes with the random utility framework of discrete choice modeling within a single end-to-end differentiable training procedure. The model has two components that are estimated together. The first is a directed dependency structure over the traveler attributes, a DAG that represents statistically supported directed pathways among traveler characteristics under observational conditional independence assumptions. The second is a discrete choice model in which the utility specification combines the traveler's observed attributes with structure-weighted interactions derived from the learned DAG. These interactions allow the discovered structure to enter the individual-level level-of-service (LOS) coefficients and alternative-specific constants (ASCs) used in the utility. The observed choice itself is excluded from the DAG and connected to the traveler attributes only through the random-utility specification, preventing the categorical outcome from distorting the structure learned over those attributes. The two objectives are optimized jointly through shared structural parameters: the structural likelihood evaluates how well the discovered DAG represents the attribute data, while the choice likelihood also informs the graph through the structure-weighted interactions used in the utility. For policy analysis, an intervention on a traveler attribute is then propagated through its downstream relationships in topological order before the utilities and choice probabilities are recomputed.

**Figure 1** illustrates this difference using a simple income intervention for a hypothetical traveler. Suppose the traveler initially has a particular combination of attributes, including income, vehicle ownership, and commuting distance. In the conventional MNL case shown, income is changed directly, while the remaining attributes are held fixed unless they are separately specified by the analyst. Neural-BSL instead propagates the same income intervention through the learned DAG, so downstream attributes such as vehicle ownership and commuting distance may also change before the utilities are recalculated. The two models therefore evaluate the same intervention under different post-intervention attribute configurations: one reflects the change in income alone, whereas the other also incorporates the downstream adjustments implied by the learned structure. These differences in the attributes entering utility can lead to different choice probabilities and, as illustrated, different mode-choice outcomes.

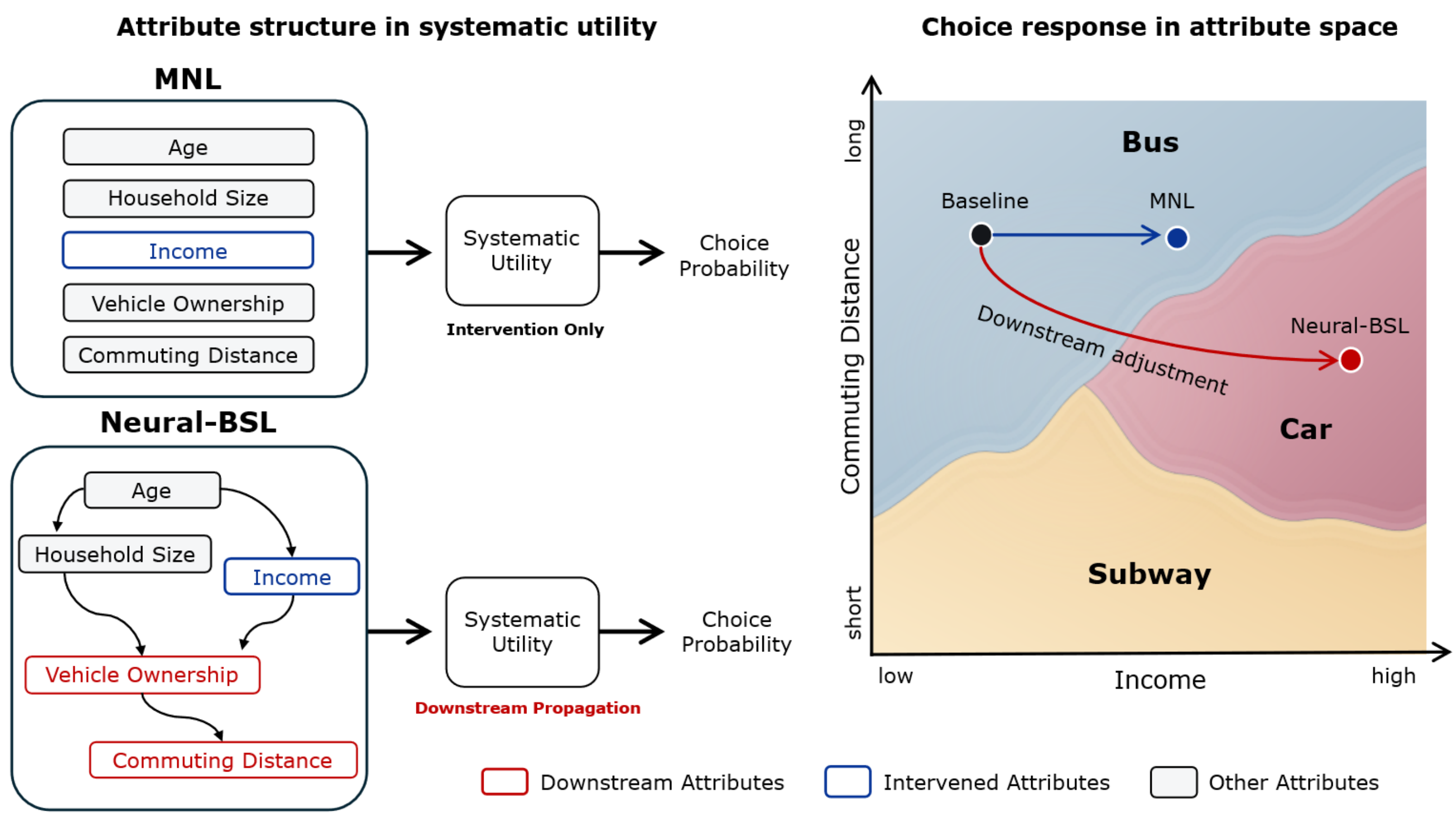

**Figure 1**. Conceptual illustration of intervention propagation and mode-choice response in MNL and Neural-BSL.

Our contributions are threefold. First, to the best of our knowledge, this is the first framework that combines structure learning and discrete choice modeling within a single end-to-end differentiable estimation procedure, allowing the discovered directed dependency structure and the fitted choice model to inform one another rather than being estimated independently, while preserving the theoretical grounding of random utility theory. Second, the framework introduces a utility specification in which individual-level LOS coefficients and ASCs depend on the observed traveler attributes together with structure-weighted interactions derived from the learned DAG. The discovered structure therefore participates directly in the fitted choice model rather than serving only as a separate descriptive output. Third, the framework supports structural policy simulations that propagate interventions through learned downstream attribute relationships before mode choice is recomputed. The resulting simulation reports not only the expected change in mode share but also the accompanying changes in downstream traveler or trip attributes, allowing the structural response underlying the forecast to be examined explicitly.

The remainder of this paper is organized as follows. **Section 2** reviews relevant literature on discrete choice modeling, structural learning, and their intersections. **Section 3** introduces the datasets. **Section 4** develops the Neural-BSL framework, including the joint training procedure and policy-simulation operator. **Section 5** presents experimental results, covering predictive performance, discovered DAGs, and policy interventions. Finally, **Section 6** concludes the paper and discusses limitations and directions for future research.

## 2. Related work

### 2.1 Random utility-based discrete choice models

The dominant tradition in travel mode choice analysis is the family of discrete choice models derived from random utility maximization (McFadden, 1974; Ben-Akiva and Lerman, 1985). The MNL model specifies the systematic utility of each alternative as a linear function of observed attributes and recovers the coefficients by maximum likelihood. Its tractability comes at the cost of the independence-of-irrelevant-alternatives property, which imposes proportional substitution across alternatives and is frequently violated in practice (Hausman and McFadden, 1984). The nested logit model relaxes this restriction by grouping correlated alternatives, and the mixed logit, or random-parameter logit, accommodates taste heterogeneity by treating selected coefficients as draws from an analyst-specified distribution (Train, 2009). A further extension, the integrated choice and latent variable (ICLV) model, introduces latent attitudinal constructs as additional explanatory variables, linking psychometric indicators to the utility through a measurement model (Ben-Akiva et al., 2002; Vij and Walker, 2016).

These models share a common interpretive strength and a common set of limitations. Their estimated coefficients carry clear behavioral meaning under the model's assumptions, which is why they remain the standard tool for policy analysis. At the same time, the systematic utility is typically linear in parameters, so any nonlinearity or interaction among attributes must be specified by the analyst in advance rather than discovered from the data. Taste heterogeneity in the mixed logit is represented through a distributional family specified by the analyst, whose parameters are then estimated from the data, and the latent constructs or structural paths in the ICLV model enter through a topology that the analyst specifies a priori rather than recovering the graph directly from data. Most relevant to the present work, observed individual attributes generally enter these choice specifications as parallel conditioning variables in the utility, while directional dependencies among the attributes themselves, such as the dependence of vehicle ownership on income or of license holding on age, are not learned from the data, leaving no internal mechanism to adjust downstream attributes when an upstream attribute is intervened upon.

### 2.2 Machine learning approaches to mode choice

A second body of work applies machine learning methods to mode choice, motivated by their ability to capture nonlinearities and interactions without explicit specification. Tree-based ensembles such as random forests and gradient-boosted trees, along with MLPs, have repeatedly been shown to match or exceed the predictive accuracy of the MNL on benchmark datasets (Hagenauer and Helbich, 2017; Lee et al., 2018; Wang and Ross, 2018; Ko et al., 2019; Kim and Bansal, 2024). The cost is interpretive: these models offer no behavioral parameters, and their predictions are explained after the fact through importance measures such as SHAP (Lundberg and Lee, 2017; Kim, 2021; Lee and Kim, 2025), which attribute a prediction to the input variables but describe little about the behavioral mechanism that produced it. A variable may therefore receive a large attribution because of its association with other variables in the observed data even when that attribution does not correspond to an intervention effect (Chauhan et al., 2025).

A more recent line of research narrows this gap by embedding neural networks inside the random utility structure rather than replacing it. The Learning Multinomial Logit model (L-MNL) divides the systematic utility into a knowledge-driven term, specified in the conventional way, and a data-driven term learned by a neural network from the remaining explanatory variables (Sifringer et al., 2020). TasteNet-MNL retains an MNL choice module but replaces fixed taste coefficients with the output of a neural network that maps individual characteristics to individual-level taste parameters, estimating the two modules jointly (Han et al., 2022). ResLogit integrates a residual neural network into the logit utility, using skip connections to capture unobserved heterogeneity while preserving the logit choice probabilities (Wong and Farooq, 2021), and related formulations couple deep networks with random utility theory to the same end (Wang et al., 2021). These hybrid models recover part of the interpretability lost in a fully black-box specification, and some impose monotonicity or sign constraints on the learned utility so that the behavioral direction of each attribute remains legible (Kim and Bansal, 2024).

What these methods do not provide is a representation of how the explanatory variables relate structurally to one another. Whether interpretation proceeds through estimated coefficients, as in the hybrid neural-utility models, or through post-hoc attribution, as in the black-box models, the resulting quantities characterize relationships under the observed data distribution rather than specifying how related attributes adjust under deliberate intervention (Brathwaite and Walker, 2018). A neural network that learns taste parameters as a function of individual characteristics such as TasteNet-MNL still treats those characteristics as parallel inputs; it does not encode that one characteristic may be upstream of another and therefore provides no mechanism for propagating an intervention through downstream attributes before choice is recomputed.

### 2.3 Structure learning

For the type of intervention analysis considered here, a directed representation of how variables may influence one another provides the structural basis for propagating a change through the covariate system. Causal modeling literature generally distinguishes between two distinct tasks. Structure learning (i.e., causal discovery) recovers the directed dependency structure among the variables, typically as a DAG, whereas causal inference estimates the magnitude of an intervention's effect once the required structural assumptions have been established (Pearl, 2009). The transportation literature has long engaged with the second task through structural equation modeling, in which the analyst specifies a recursive ordering among travel-related variables by hand and estimates the path coefficients along the assumed structure (Golob, 2003). The structure in this tradition is supplied by theory and hand-crafted a prior rather than recovered from the data, and its credibility rests on the plausibility of the analyst's specification.

Structure learning instead addresses the first task by recovering the DAG from the data rather than assuming it in advance. Constraint-based algorithms such as the Peter-Clark algorithm orient edges using conditional independence tests (Spirtes and Glymour, 1991), while score-based methods such as greedy equivalence search navigate the space of graphs by optimizing a likelihood-based score (Spirtes et al., 2000; Chickering, 2003). Both rely on combinatorial search over a space that grows super-exponentially in the number of variables (Robinson, 1977). The reformulation introduced by NOTEARS (Zheng et al., 2018) and the Bayesian extension developed in DiBS (Lorch et al., 2021) replace this search with gradient-based optimization over a smooth representation of the graph. This continuous formulation, which we refer to as differentiable structure learning, is what makes structure learning compatible with the gradient-based estimation of a choice model.

A small but growing literature has begun to bring these tools to bear on travel mode choice. Brathwaite and Walker (2018) drew attention to the general disconnect between travel demand modeling and causal modeling and set out a conceptual case for causal mode choice models. On the empirical side, Xie and Waller (2010) learned a Bayesian network for discrete travel choice from observational data and prior hypotheses, and Ma et al. (2017) combined structural restrictions with model averaging over Bayesian networks. Most directly related to the present work, Chauhan et al. (2024) applied constraint-based and score-based discovery algorithms to mode choice and fed the recovered graph into a structural equation model to quantify structural relationships, establishing that a data-driven dependency structure over traveler attributes can provide behaviorally informative results. The authors note several limitations that motivate the present study: the LOS attributes of the alternatives were excluded for data reasons, the structural relations were restricted to linear forms, and unobserved confounding was left for future work.

Two methodological limitations are shared across these studies. First, the observed choice is represented as a node within the graph rather than retained as the outcome of a comparison among alternative-specific utilities. Such a representation does not preserve the random utility mechanism through which alternative-specific LOS attributes enter the choice process and introduces artificial structural dependencies due to the mutually exclusive encoding of choice outcomes. Second, the dependency structure is learned in a stage separate from the choice model, so the structural score and the choice likelihood cannot jointly inform the graph parameters. The recovered structure can therefore describe relationships among attributes, but it is not integrated into the utility model as a mechanism for propagating an intervention through downstream attributes to mode choice.

Across these literatures, no existing framework recovers the dependency structure among individual attributes and estimates a discrete choice model within a single differentiable procedure. The discrete choice and neural-utility models of **Sections 2.1** and **2.2** do not learn a directed structure among the observed individual attributes. The structural equation tradition specifies those relationships in advance rather than discovering them, while the structure learning studies above estimate the graph separately from the choice model without preserving alternative-specific random utility specifications. The framework proposed in this paper addresses this gap by jointly learning the directed dependency structure over the individual attributes and the choice model under a unified differentiable objective. The discovered structure enters the choice model through the random utility specification and is used to trace how an intervention propagates through downstream attributes to mode-choice probabilities.

## 3. Dataset

We adopt two complementary mode choice datasets that differ in scale, geography, and elicitation method: a stated-preference (SP) survey conducted in the Seoul Metropolitan Area, and the revealed-preference (RP) London Passenger Mode Choice (LPMC) dataset of Hillel et al. (2018). The two datasets the framework across distinct data-generating environments. The SP data provide a designed choice experiment, in which the LOS attributes are exogenously assigned by the experimental design, with a set of attitudinal items and a sample size typical of SP studies. The LPMC data complement the SP data by evaluating the framework on large-scale revealed behavior, where the LOS attributes represent observed, network-derived supply conditions rather than experimentally assigned. Evaluating both environments validates a key specification of our framework: treating LOS attributes as exogenous supply outside the traveler-attribute DAG, under both controlled experimental designs and real-world network conditions. Descriptive statistics for both datasets are summarized in **Tables 1** and **2**; preprocessing steps common to both are described in **Section 3.3**.

### 3.1 Stated-preference dataset (Seoul Metropolitan Area)

The SP dataset was collected via a web-based survey of commuters in the Seoul Metropolitan Area between October and December 2024 by Park and Kim (2025), to which the reader is referred for full details of the survey design and recruitment. The final sample comprised 863 respondents, each completing six choice tasks, yielding 5,178 observations. Each task presented a commuting mode-choice scenario with three alternatives: bus (BUS), subway (SUB), and demand-responsive transit (DRT). The scenarios were constructed using a D-efficient design, with variation in five LOS attributes: in-vehicle time, walking access time, waiting time, transfers, and fare (Bliemer and Rose, 2009). Bus and subway in-vehicle time and fare were held constant across scenarios by design and were therefore excluded from estimation, leaving eleven varying mode-specific LOS attributes in the utility specification. The realized mode shares were 19.1% BUS, 29.6% SUB, and 51.2% DRT.

Each respondent is described by seventeen individual-level attributes: eight binary sociodemographic indicators, five commuting-pattern variables, and four continuous attitudinal factor scores. The attitudinal factor scores are derived from seventeen five-point Likert attitudinal items covering four constructs (punctuality, efficiency, comfort, innovativeness) via minimum-residual factor analysis with varimax rotation on the training respondents; loadings and goodness-of-fit are reported in **Appendix A**. We split the dataset 70/15/15 by respondent identifier into 604 training, 129 validation, and 130 test respondents (3,624 / 774 / 780 scenario-level observations), ensuring that all six scenarios from a given respondent fall into the same partition.

**Table 1**. Descriptive statistics for the SP dataset (N = 863 respondents; 5,178 scenario-level observations).

| Variable | Code | Type | Mean / Prop. | SD / Range |
|---|---|---|---|---|
| ***Mode share (5,178 scenarios)*** | | | | |
| Bus / Subway / DRT | BUS / SUB / DRT | Proportion | 19.1 / 29.6 / 51.2 % | |
| ***Sociodemographics (respondent-level)*** | | | | |
| Male | male | Binary | 53.4 % | |
| Age 50 or above | age_50 | Binary | 25.4 % | |
| Education at or below high school | edu_high | Binary | 16.7 % | |
| Monthly income at or above 3M KRW | wage_300 | Binary | 60.5 % | |
| Holds a driver's license | LICENSE | Binary | 85.9 % | |
| Owns a car | CAR | Binary | 71.8 % | |
| Professional or managerial occupation | job_pro | Binary | 19.8 % | |
| Production or technical occupation | job_gen | Binary | 7.0 % | |
| ***Commuting patterns (respondent-level)*** | | | | |
| Transfers on the respondent's usual commute | transfers_total | Count | 1.5 | sd 1.01, [0,3] |
| Walking access time at or above 15 min | walk_15min | Binary | 35.3 % | |
| Waiting time at or above 15 min | wait_15min | Binary | 33.8 % | |
| Commutes 5 or more days per week | day_5 | Binary | 85.5 % | |
| Commute time at or above 60 min | comm_1 | Binary | 57.2 % | |
| ***Attitudinal items (five-point Likert scale, 1 to 5; item-level statistics, see Appendix A)*** | | | | |
| Punctuality items (P_1 to P_5) | F_punct (factor) | 5 items | mean 3.05 to 4.31 | sd 0.70 to 1.00 |
| Efficiency items (E_1 to E_4) | F_effic (factor) | 4 items | mean 3.73 to 4.26 | sd 0.70 to 0.93 |
| Comfort items (C_1 to C_4) | F_comf (factor) | 4 items | mean 3.40 to 4.25 | sd 0.73 to 1.05 |
| Innovativeness items (N_1 to N_4) | F_innov (factor) | 4 items | mean 3.64 to 4.09 | sd 0.77 to 0.91 |
| ***Level-of-service attributes (scenario-level)*** | | | | |
| Bus walking access time | BUS_ACC | Discrete, min | 12.3 | [10, 15] |
| Bus waiting time | BUS_WAIT | Discrete, min | 14.2 | [10, 20] |
| Bus transfers | BUS_TRANSFER | Binary | 0.33 | |
| Subway walking access time | SUB_ACC | Discrete, min | 15.6 | [10, 20] |
| Subway waiting time | SUB_WAIT | Discrete, min | 5.6 | [3, 8] |
| Subway transfers | SUB_TRANSFER | Binary | 0.58 | |
| DRT in-vehicle time | DRT_TT | Discrete, min | 54.6 | [45, 65] |
| DRT walking access time | DRT_ACC | Discrete, min | 7.5 | [5, 10] |
| DRT waiting time | DRT_WAIT | Discrete, min | 7.5 | [5, 10] |
| DRT transfers | DRT_TRANSFER | Binary | 0.50 | |
| DRT fare | DRT_FARE | Discrete, KRW | 3,500 | [3,000, 4,000] |

### 3.2 Revealed-preference dataset (LPMC)

The LPMC dataset of Hillel et al. (2018) is a large-scale benchmark constructed by linking three years of trip-diary records from the London Travel Demand Survey (April 2012 to March 2015) with mode-specific LOS attributes computed from an online journey planner and a tailored cost model for the unchosen alternatives. The dataset contains 81,086 trips made by 31,954 individuals across 17,616 households, with the choice set comprising walking (17.6% mode share), cycling (3.0%), public transport (35.3%), and driving (44.2%).

Following the description of the LPMC dataset, we use six person-level attributes: gender, age, driving-license possession, an indicator for any car in the household, and two fare-concession indicators, one for the age-based entitlements and one for the disabled entitlement. The raw fare field distinguishes five types, of which three are granted on age (child, 16-plus, and the over-60 pass) and one on disability; because the age-based types are held at both ends of the age range while the disabled type is not, we keep the two entitlements as separate nodes rather than as a single count. These are used together with five trip-context features that we derive from the raw fields: log-transformed straight-line trip distance, a commute-purpose indicator (home-based work or education), a rush-hour indicator (07-09 or 17-19), a weekend indicator, and a winter indicator. Trip-context features are treated as individual covariates in the structural attribute DAG because they describe the exogenous conditions under which the trip is made, not the supply of any alternative. The LOS block contains twelve mode-specific attributes; we retain eleven after the multicollinearity correction described in **Section 3.3**. We split the dataset grouped by household identifier into 56,493 training, 12,414 validation, and 12,179 test trips to prevent intra-household data leakage. For structure learning we use a subset of one randomly sampled trip per training individual (22,282 rows), which removes the repeated-trip dependence within travelers, while choice prediction is trained on all 56,493 training trips.

**Table 2**. Descriptive statistics for the LPMC (RP) dataset (N = 81,086 trips; 17,616 households; 31,954 individuals). Statistics are computed at trip level.

| Variable | Code | Type | Mean / Prop. | SD / Range |
|---|---|---|---|---|
| ***Mode share (81,086 trips)*** | | | | |
| Walk / Cycle / PT / Drive | walk / cycle / PT / drive | Proportion | 17.6 / 3.0 / 35.3 / 44.2 % | |
| ***Person attributes*** | | | | |
| Female | female | Binary | 52.6 % | |
| Age | age | Continuous, years | 39.5 | sd 19.2, [5, 99] |
| Holds a driver's license | has_license | Binary | 61.7 % | |
| Any car in the household | car_any | Binary | 70.8 % | |
| Age-based fare concession | conc_age | Binary | 32.5 % | |
| Disabled fare concession | conc_disab | Binary | 4.8 % | |
| ***Trip context*** | | | | |
| Straight-line trip distance | distance_log | Continuous, m | 4,605 | sd 4,782, [77,40,941] |
| Commute trip (work or education) | is_commute | Binary | 28.0 % | |
| Rush hour (07 to 09 or 17 to 19) | rush_hour | Binary | 28.7 % | |
| Weekend trip | weekend | Binary | 25.9 % | |
| Winter trip (Dec to Feb) | winter | Binary | 22.7 % | |
| ***Level-of-service attributes (after corrections; see Section 3.3)*** | | | | |
| Walking duration | dur_walking | Continuous, hr | 1.13 | sd 1.12, [0.03, 9.28] |
| Cycling duration | dur_cycling | Continuous, hr | 0.36 | sd 0.35, [0.01, 3.05] |
| PT access and egress duration | dur_pt_access | Continuous, hr | 0.16 | sd 0.09, [0, 1.19] |

| PT rail in-vehicle duration | dur_pt_rail | Continuous, hr | 0.09 | sd 0.18, [0, 1.47] |
|---|---|---|---|---|
| PT bus in-vehicle duration | dur_pt_bus | Continuous, hr | 0.17 | sd 0.19, [0, 2.15] |
| PT interchange duration | dur_pt_int | Continuous, hr | 0.04 | sd 0.08, [0, 0.87] |
| Number of PT interchanges | pt_interchanges | Count | 0.37 | sd 0.62, [0, 4] |
| PT fare per straight-line km | pt_fare_per_km | Continuous, GBP/km | 0.61 | sd 0.89, [0, 15.00] |
| Driving duration | dur_driving | Continuous, hr | 0.28 | sd 0.25, [0, 2.06] |
| Driving congestion charge | cost_driving_ccharge | Continuous, GBP | 1.07 | sd 3.18, [0, 10.50] |
| Predicted driving traffic variability | driving_traffic_percent | Continuous, [0, 1] | 0.34 | sd 0.20, [0, 1.25] |

### 3.3 Common preprocessing

The two datasets are preprocessed under a common pipeline before being passed to the framework. Attributes entering the utility are standardized to zero mean and unit variance using training-set statistics only. The structural score of **Section 4.2** instead takes binary indicators on their native 0/1 scale and continuous attributes standardized using training-set statistics, consistent with the Bernoulli and unit-variance Gaussian likelihoods used for the respective node types. Choices are encoded as integer labels to evaluate the multinomial negative log-likelihood loss.

The LPMC LOS block contains two cost variables with multicollinearity that require correction before they can be used as independent predictors. The driving fuel cost has a variance-inflation factor of 24 against the remaining LOS variables and a correlation of 0.97 with raw trip distance, reflecting that the fuel-cost field is largely determined by trip length in the LPMC generation process (Hillel et al., 2018) and provides little independent variation. The total transit fare is moderately collinear with several distance-driven variables, since the fare structure scales with distance. We discard driving fuel cost permanently to eliminate deterministic collinearity and replace transit fare with a per-kilometer fare rate, computed as transit fare divided by straight-line trip distance in kilometers (with distances floored at 100 m to avoid division by near-zero values, which affects nine trips). After this correction, the LPMC LOS block comprises eleven mode-specific attributes, with the most severe distance-driven multicollinearity removed.

## 4. Methodology

This section develops the Neural-BSL framework. We first introduce the notation and provide an architectural overview (**Section 4.1**), then describe the differentiable structure learning module (**Section 4.2**), the utility specification that connects the discovered structure to the choice probabilities (**Section 4.3**), and finally the joint training procedure together with the policy-simulation (intervention) operator (**Section 4.4**).

### 4.1 Problem setup and overview

We consider a discrete choice setting in which each choice observation $n \in \{1, \ldots, N\}$ records a traveler facing a choice set $\mathcal{J}$ consisting of $J$ alternatives and selecting one alternative $y_n \in \mathcal{J}$. The data describing each observation (a single stated-preference scenario or an observed trip) are organized into two groups of variables that play structurally different roles.

The first group, the individual attributes $\boldsymbol{X}_n \in \mathbb{R}^{D_X}$, describes characteristics of the traveler such as age, income, gender, and license holding, encoded as real values as described in **Section 3**. Throughout, $n$ indexes choice observations, $i$ and $r$ index individual attributes, $j$ indexes choice alternatives, and $k$ indexes particles. The observed choice $y_n$ is excluded from the DAG and is connected to $\boldsymbol{X}_n$ only through the random-utility specification. Accordingly, structure learning (**Section 4.2**), the structure-weighted interaction (**Section 4.3**), and policy interventions on individual attributes (**Section 4.4**) operate exclusively on the $D_X \times D_X$ graph over $\boldsymbol{X}$.

The second group is the LOS vector $\boldsymbol{g}_n \in \mathbb{R}^{D_G}$, which contains alternative-specific quantities such as in-vehicle time, access time, and fare. Here, $D_G$ is the total number of scalar LOS attributes, $\ell \in \{1, \dots, D_G\}$ indexes them and $\mathcal{L}_j \subseteq \{1, \dots, D_G\}$ denotes the subset entering the utility of alternative $j$. LOS attributes describe what each alternative offers under a given supply configuration and enter the utility directly, without passing through the attribute DAG.

The objective is twofold. First, we recover a directional dependency structure over $\boldsymbol{X}$ that encodes the directed relationships among individual attributes under observational conditional independence assumptions. Second, conditional on this structure, we estimate a discrete choice model whose individual-level utility reflects the discovered structure, so that interventions on individual attributes propagate through the discovered edges and affect both choice probabilities and the values of other individual attributes.

Neural-BSL is organized into three blocks, of which the first two carry the trainable parameters and are estimated jointly under a single objective: (i) differentiable structure learning, which returns a posterior distribution over DAGs on $\boldsymbol{X}$; (ii) utility specification, in which the discovered structure enters through an interaction term $\boldsymbol{s}_{n,k}$ that pairs each attribute with the weighted sum of its parents, and two networks produce the LOS coefficients and ASCs; and (iii) discrete choice, in which these combine with the observed LOS attributes to form the random utility of each alternative, with the choice probabilities following from a softmax operator. To represent structural uncertainty, all parameters are maintained in $K$ parallel particles indexed by $k \in \{1, \dots, K\}$, with each particle realizing a specific configuration of the graph structure and its associated utility parameters (i.e., a distinct sample from the joint posterior distribution); the particles together approximate the posterior distribution over models given the data. The linear utility components of Neural-BSL are warm-started from a specification-matched MNL baseline, while its additional graph-dependent and nonlinear components are initialized at small magnitude. **Figure 2** summarizes the framework, comprising differentiable structure learning and utility specification in the upper panels and discrete choice in the lower panel.

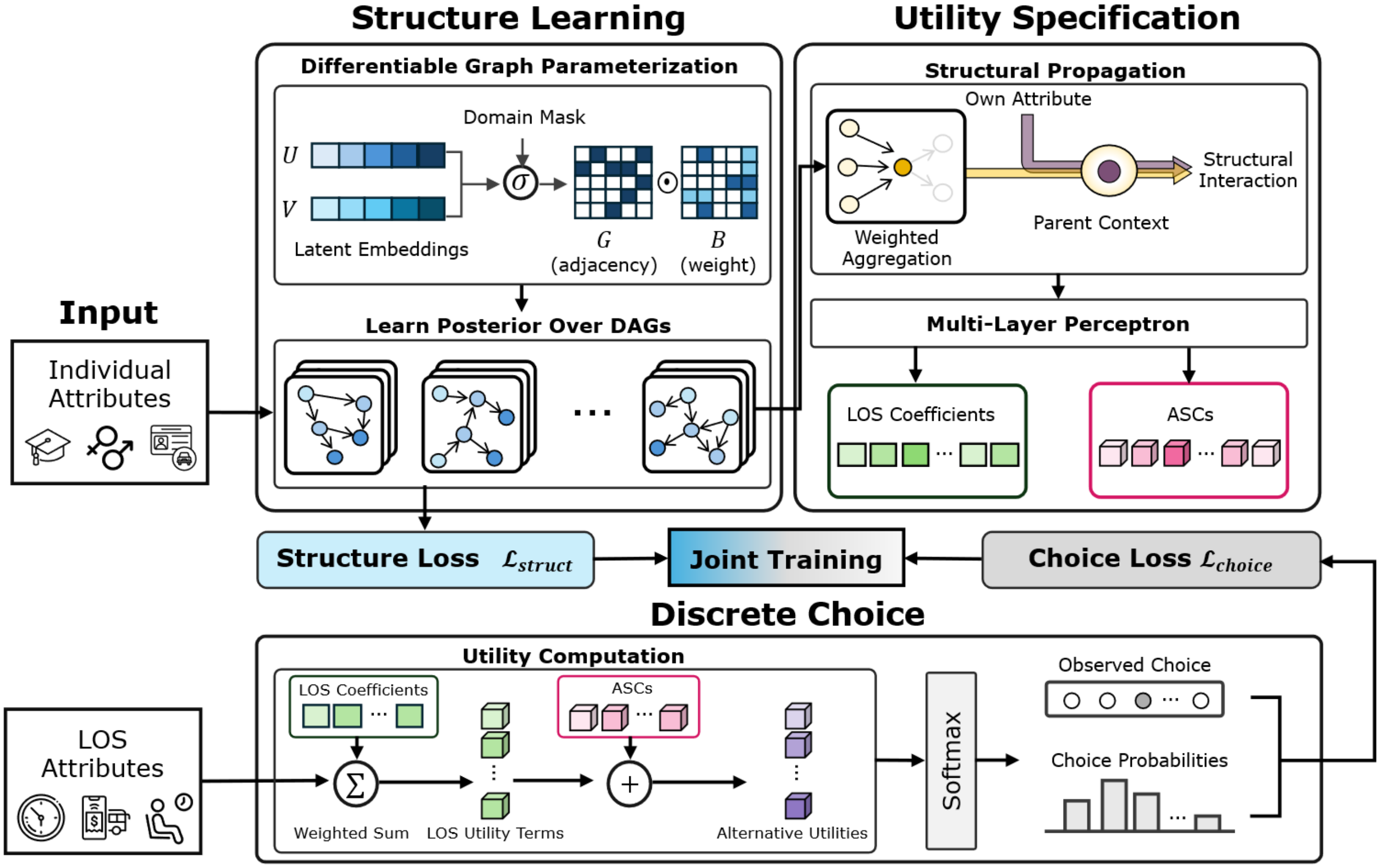

**Figure 2**. Overall Architecture of the Neural-BSL framework

### 4.2 Differentiable structure learning

The directed dependency structure over the $D_X$ individual attributes is represented as a DAG over the variables in $\boldsymbol{X}$. We learn a posterior distribution over such graphs through $K$ parallel particles, each parameterizing a soft adjacency matrix $\boldsymbol{G}_k \in [0,1]^{D_X \times D_X}$ whose entries represent the probabilities of directed edges. Recovering this structure from observational data is challenging for three reasons. First, the number of possible DAGs grows super-exponentially with $D_X$, making exhaustive search intractable beyond a few variables. Second, observational travel mode choice datasets often provide limited information for uniquely determining the dependency structure, which can induce substantial structural ambiguity and render deterministic point-estimate graphs susceptible to overfitting or recovering spurious edges. Third, a single point-estimate graph offers no measure of confidence in the discovered edges, which is critical when the structure is used downstream for policy simulation: an edge supported by overwhelming evidence should not be treated identically to one whose existence is uncertain.

We adopt the differentiable Bayesian structure learning framework (Lorch et al., 2021), which addresses these limitations by recasting structure learning as a continuous optimization problem and returning a posterior distribution over graphs rather than a single estimate. By maintaining a particle-based variational approximation of the posterior over DAGs, our framework explicitly captures structural uncertainty arising from limited observational information, preventing the framework from overcommitting to a single fragile topology.

The idea is to give each attribute a low-dimensional latent embedding and to let the alignment between two latent vectors determine how likely an edge is. Searching over discrete graphs then becomes optimization over continuous vectors (i.e., smooth embeddings). Each candidate graph is represented through a pair of low-dimensional latent matrices $\boldsymbol{Z}_k = (\boldsymbol{U}_k, \boldsymbol{V}_k) \in \mathbb{R}^{D_X \times d} \times \mathbb{R}^{D_X \times d}$, where $d \ll D_X$ is the embedding dimension. The entry $\boldsymbol{G}_k[i,r]$ represents the probability of the directed edge $r \to i$; thus, rows index children and columns index parents. The probability that node $r$ is a parent of node $i$ under particle $k$ is expressed as **Equation 1**.

$$\boldsymbol{G}_k[i,r] = \sigma\left(\frac{\left(\boldsymbol{U}_k \boldsymbol{V}_k^{\top}\right)[r,i]}{\tau} + \boldsymbol{M}[i,r]\right) \cdot \mathbb{1}[i \neq r] \tag{1}$$

where $\sigma(\cdot)$ is the sigmoid function, $\tau > 0$ is a temperature controlling edge sharpness, $\boldsymbol{M} \in \{-\infty, 0\}^{D_X \times D_X}$ is a logit-space mask with $\boldsymbol{M}[i,r] = 0$ for allowed edges and $\boldsymbol{M}[i,r] = -\infty$ for forbidden edges, and the indicator $\mathbb{1}[i \neq r]$ rules out self-loops. The mask rules out edges that cannot exist on prior grounds, such as any edge pointing into gender; the complete specification of $\boldsymbol{M}$ for each dataset is provided in **Appendix B**. The temperature parameter $\tau$ provides a continuous relaxation of the discrete graph search space: higher initial values yield a smooth adjacency matrix $\boldsymbol{G}_k$ necessary for gradient-based exploration, whereas systematically annealing $\tau \to 0$ transitions $\boldsymbol{G}_k$ into a deterministic, near-binary structure as optimization proceeds. Crucially, the smooth differentiability of $\boldsymbol{G}_k$ with respect to the latent embeddings $(\boldsymbol{U}_k, \boldsymbol{V}_k)$ enables end-to-end gradient flow, allowing the downstream choice objective (**Section 4.3**) to jointly guide structural optimization alongside the structural likelihood.

A DAG, by definition, contains no directed cycles. Because cycles correspond to combinatorial constraints on the adjacency matrix, they cannot be imposed directly during gradient-based optimization. Yu et al. (2019) addressed this by showing that the matrix function:

$$h(\boldsymbol{G}_k) = \operatorname{tr}((\boldsymbol{I} + \boldsymbol{G}_k / D_X)^{D_X}) - D_X \tag{2}$$

is non-negative for any matrix with non-negative entries (which $\boldsymbol{G}_k \in [0,1]^{D_X \times D_X}$ satisfies by construction) and equals zero if and only if $\boldsymbol{G}_k$ encodes an acyclic graph; intuitively, $h(\cdot)$ counts, in a smoothed sense, the total weight of directed cycles of all lengths through powers of the adjacency matrix. Adding $h(\boldsymbol{G}_k)$ to the training loss under an augmented-Lagrangian schedule (**Section 4.4**) thus converts acyclicity from a hard combinatorial constraint into a differentiable penalty.

To assess how well a candidate graph explains the observed attribute data, we adopt a mixed-type generalized linear score, in which each attribute is modeled conditional on its parents by the distribution appropriate to its measurement scale. The score here serves as a functional-form assumption on how variables relate to their parents once the parent set is fixed. This is the same linear-additive parents-to-child specification that underlies the structural equation models routinely used in transportation research. However, in our framework, structural relations are precisely what the framework learns from data through $(\boldsymbol{U}_k, \boldsymbol{V}_k)$, which distinguishes our use of structural equations from confirmatory frameworks such as the ICLV model, in which the analyst specifies the graph topology a priori and uses the data only to fit its magnitudes. In addition to $(\boldsymbol{U}_k, \boldsymbol{V}_k)$, each particle maintains a structural regression matrix $\boldsymbol{B}_k \in \mathbb{R}^{D_X \times D_X}$ encoding candidate edge weights (analogous to path parameters in structural equation modeling), together with a vector of node intercepts $\boldsymbol{b}_k = \left(b_{1,k}, \dots, b_{D_X,k}\right)^\top$. For structural learning, $m \in \{1, \dots, N_{struct}\}$ indexes the $N_{struct}$ independent attribute observations $\boldsymbol{X}_m \in \mathbb{R}^{D_X}$ in the structure-learning dataset. For observation $m$, each attribute $X_{m,i}$ is modeled through a structural generalized linear predictor $\eta_{m,i,k}$ formed from its parent attributes in the discovered graph, as shown in **Equation 3**.

$$\eta_{m,i,k} = b_{i,k} + \sum_{r=1}^{D_X} (\boldsymbol{G}_k \odot \boldsymbol{B}_k)[i,r]\, X_{m,r} \tag{3}$$

where $b_{i,k}$ is the $i$-th element of the structural intercept vector $\boldsymbol{b}_k$. The sum formally runs over all other nodes: edges excluded by the mask $\boldsymbol{M}$ contribute exactly zero, and the remaining candidate parent attributes are weighted by the soft edge probability $\boldsymbol{G}_k$. The $\boldsymbol{B}_k$ specifies the underlying magnitude and sign of potential direct effects within the generalized linear predictor, while $\boldsymbol{G}_k$ serves as a probabilistic edge-existence gate. The fitted structural coefficient governing the influence of parent $r$ on child $i$ is therefore the elementwise product $(\boldsymbol{G}_k \odot \boldsymbol{B}_k)[i,r]$, rather than $\boldsymbol{B}_k[i,r]$ alone.

Let $f_i(x \mid \eta)$ denote the conditional probability mass function for a binary node and the conditional density for a continuous node. The fitted mixed-type conditional is:

$$f_i(x \mid \eta) = \begin{cases} \text{Bernoulli}\left(x; \sigma(\eta)\right), & i \text{ is binary,} \\ \mathcal{N}(x; \eta, 1), & i \text{ is Gaussian.} \end{cases} \tag{4}$$

Binary attributes enter on their native 0/1 scale. Continuous attributes enter on the scale established by the training-data preprocessing, with the unit Gaussian variance corresponding to the standardized continuous variables.

No separate prior is placed on $\boldsymbol{B}_k$; it enters the structural likelihood through $\boldsymbol{G}_k \odot \boldsymbol{B}_k$, while sparsity is imposed on $\boldsymbol{G}_k$. While observational score functions alone cannot uniquely orient edges among statistically equivalent graphs (i.e., a Markov equivalence class), the prior plausibility mask $\boldsymbol{M}$ together with the downstream choice likelihood (**Section 4.3**) provide structural identification constraints, with posterior uncertainty among admissible graphs represented across the ensemble particles. The structural loss is the negative log-likelihood, defined as:

$$\mathcal{L}_{struct,k} = -\sum_{m=1}^{N_{struct}} \sum_{i=1}^{D_X} log\, f_i\left(X_{m,i} \mid \eta_{m,i,k}\right). \tag{5}$$

**Equation 5** evaluates both the plausibility of the structure $\boldsymbol{G}_k$ and the fit of the magnitudes $\boldsymbol{B}_k$ to the structural data. We deliberately retain $\boldsymbol{B}_k$ as an explicit trainable quantity rather than integrating it out under a closed-form score such as the Bayesian-Gaussian-equivalent score (Kuipers and Moffa, 2025), because the structure-weighted interaction (**Section 4.3**) and the topological policy simulations (**Section 4.4**) require explicitly fitted structural magnitudes. A score function that marginalizes $\boldsymbol{B}_k$ would require a separate post-hoc regression step on the discovered graph, breaking the joint differentiability that unifies structure learning and choice estimation.

The $K$ particle-specific structural parameter sets $\{(\boldsymbol{Z}_k, \boldsymbol{B}_k, \boldsymbol{b}_k)\}_{k=1}^{K}$, together with their utility-model parameters, approximate the posterior over the complete joint model. They are updated jointly with the choice-model parameters through Stein Variational Gradient Descent (SVGD; Liu and Wang, 2016), in which each particle is attracted toward regions of high posterior density while a kernel-based repulsive term prevents particle collapse. The full update rule is given in **Section 4.4**.

### 4.3 Utility specification

Given a candidate directed dependency structure encoded by $\boldsymbol{G}_k$ and $\boldsymbol{B}_k$, the model proceeds in three steps: the discovered weights are rescaled and combined with the individual attributes to form the structure-weighted interaction (**Equations 6-7)**; the neural components generate the individual-level LOS coefficients and ASCs (**Equations 8-9**), and the observed LOS attributes enter the utility directly, with choice probabilities obtained through the softmax function (**Equations 10-11**). The effective structural weight matrix $\boldsymbol{A}_k = \boldsymbol{G}_k \odot \boldsymbol{B}_k \in \mathbb{R}^{D_X \times D_X}$ combines the existence of each edge with its magnitude. Because a Bernoulli node carries its coefficient on the logit scale while a Gaussian node carries it on the standardized scale, the entries of this matrix are not directly comparable across nodes. We therefore rescale every entry to a common standard-deviation-per-standard-deviation footing (**Equation 6):**

$$\boldsymbol{A}_{std,k}[i,r] = c_i \cdot \boldsymbol{A}_k[i,r] \cdot \mathrm{sd}(X_r) \tag{6}$$

where the row factor $c_i$ is $\sqrt{p_i(1-p_i)}$ when child node $i$ is binary (where $p_i$ is the sample proportion) and $1/sd(X_i)$ when continuous, and $\mathrm{sd}(X_r)$ is the sample standard deviation of parent attribute $r$. This is a change of units, not of the fitted structure, and $\boldsymbol{A}_{std,k}$ is the matrix used from here on. The structure reaches the utility through one quantity, the elementwise product of a traveler's attributes with the weighted sum of their parents:

$$\boldsymbol{s}_{n,k} = \boldsymbol{X}_n \odot \left(\boldsymbol{A}_{std,k}\, \boldsymbol{X}_n\right) \tag{7}$$

so that the $i$-th coordinate of $\boldsymbol{s}_{n,k}$ pairs the value of node $i$ with the weighted sum of its parents under $\boldsymbol{A}_{std,k}$. The interaction is one step deep: it carries a node with its immediate parents, not its full ancestry. As an illustration, a traveler whose license-holding indicator equals one and whose license-holding is structurally supported, through the discovered structure, by gender, age, and education, receives a license-holding coordinate whose magnitude reflects the strength of those upstream links. Through this structure-weighted interaction, the choice-prediction loss gradient $\nabla_{s_{n,k}}\mathcal{L}_{choice,k}$ (**Equation 12** below) backpropagates directly into $\boldsymbol{G}_k$ and $\boldsymbol{B}_k$ via the chain rule, enabling end-to-end joint estimation.

The LOS coefficients and the ASCs are generated from the same input, the concatenation $\boldsymbol{z}_{n,k} = \boldsymbol{X}_n \oplus \boldsymbol{s}_{n,k} \in \mathbb{R}^{2D_X}$. This vector is the input to the nonlinear component of both the LOS coefficients and the ASCs. The two blocks serve distinct roles: $\boldsymbol{X}_n$ carries the linear effect of directly observed characteristics, and $\boldsymbol{s}_{n,k}$ carries, for each characteristic, its interaction with the causal context the discovered graph assigns to it. In the language of discrete choice modeling, the first block corresponds to a standard linear utility term and the second to an interaction term whose weights are the discovered structural coefficients rather than products chosen by the analyst.

The coefficients on the LOS attributes of traveler $n$ under particle $k$ are produced according to:

$$\boldsymbol{\beta}_{n,k} = \boldsymbol{\beta}_{base,k} + \boldsymbol{\gamma}_k \boldsymbol{X}_n + \boldsymbol{\gamma}_{pair,k}\, \boldsymbol{v}_n + \boldsymbol{\gamma}_{s,k}\, \boldsymbol{s}_{n,k} - softplus\left(m_k(\boldsymbol{z}_{n,k})\right) + log\ 2 \tag{8}$$

where $\boldsymbol{\beta}_{base,k} \in \mathbb{R}^{D_G}$ is a population-level coefficient vector, $\boldsymbol{\gamma}_k \in \mathbb{R}^{D_G \times D_X}$ is a linear coefficient matrix capturing interactions between the LOS attributes and the individual attributes and $\boldsymbol{\gamma}_{pair,k} \in \mathbb{R}^{D_G \times P}$ is a coefficient matrix on a small, fixed set of $P$ pairwise products of selected individual attributes (e.g., license × car ownership), collected in the vector $\boldsymbol{v}_n \in \mathbb{R}^P$; the same pairwise terms appear in the MNL baseline. These three terms replicate the standard specification-matched MNL baseline.

The matrix $\boldsymbol{\gamma}_{s,k} \in \mathbb{R}^{D_G \times D_X}$ carries the structure-weighted interaction into the LOS coefficients, allowing the learned attribute DAG topology ($\boldsymbol{s}_{n,k}$) to directly shift the LOS sensitivities. The mapping $m_k: \mathbb{R}^{2D_X} \rightarrow \mathbb{R}^{D_G}$ is a small MLP (one hidden layer of width 16, tanh activation) producing the nonlinear residual driven by both direct attributes and structural contexts. The offset formulation $-softplus\left(m_k(\boldsymbol{z}_{n,k})\right) + log\ 2$ centers the nonlinear residual at zero when $m_k(\boldsymbol{z}_{n,k}) = 0$, so that near-zero initialization of the output layer keeps the model close to the specification-matched MNL baseline at the start of training. The parameters $\boldsymbol{\beta}_{base,k}$ and $\boldsymbol{\gamma}_k$ are warm-started from the maximum-likelihood estimates of a specification-matched MNL baseline that shares the identical variable list and the same linear interaction structure; the output layer of $m_k$ is initialized near zero. The coefficients multiplying $\boldsymbol{s}_{n,k}$ are initialized at zero, while the nonlinear output layers are initialized at small magnitude. This design ensures that Neural-BSL strictly generalizes the conventional MNL baseline, starting from the baseline solution while smoothly learning structure-weighted and nonlinear behavioral refinements during training.

The specification extends the matched MNL through the structure-weighted and nonlinear terms while retaining the same linear and pairwise interaction components. Because the additive specification may produce a positive LOS coefficient, the total training objective includes a penalty on positive coefficients (**Equation 13**), where ${\beta_{n,k}}^{(\ell)}$ denotes the coefficient on LOS dimension $\ell$.

The ASCs are produced from the same input $\boldsymbol{z}_{n,k}$ and follow a parallel additive form:

$$ASC_{n,j,k} = ASC_{base,j,k} + \boldsymbol{\gamma}_{asc,k,j} \boldsymbol{X}_n + \boldsymbol{\gamma}_{s,asc,k,j} \boldsymbol{s}_{n,k} + \delta_{asc,k,j}(\boldsymbol{z}_{n,k}). \tag{9}$$

where $ASC_{base,j,k}$ is a population-level constant, $\boldsymbol{\gamma}_{asc,k,j}$ is the linear coefficient vector applied to $\boldsymbol{X}_n$, $\boldsymbol{\gamma}_{s,asc,k,j}$ is the linear coefficient vector capturing the structural interaction $\boldsymbol{s}_{n,k}$, and the nonlinear term $\delta_{asc,k,j}(\boldsymbol{z}_{n,k})$ is implemented as the sum of two parallel MLP heads, each with one hidden layer of eight tanh units. The reference alternative (in our applications, bus for the SP dataset and walking for the RP dataset) has its ASC fixed to zero, following the standard identification convention in random utility models.

The LOS coefficients and ASCs combine with the observed LOS attributes through a separate pathway that does not pass through the attribute DAG. Embedding LOS values in the DAG would conflate the experimental design (in SP data, LOS values are assigned exogenously by the experimenter) or the network-determined supply (in RP data) with the dependency structure of individual attributes, which is conceptually distinct. The observed vector of LOS attributes $\boldsymbol{g}_n \in \mathbb{R}^{D_G}$ therefore enters the utility as they stand, without gating or rescaling by the structural component. The utility of alternative $j$ for traveler $n$ under particle $k$ is then defined as **Equation 10**.

$$V_{n,j,k} = ASC_{n,j,k} + \sum_{\ell \in \mathcal{L}_j} \beta_{n,k}{}^{(\ell)} \cdot g_n{}^{(\ell)} \tag{10}$$

where $g_n{}^{(\ell)}$ denotes the observed scalar value of the $\ell$-th LOS attribute for traveler $n$, and the sum runs over the set of LOS indices $\mathcal{L}_j \subseteq \{1, \dots, D_G\}$ entering the systematic utility of alternative $j$. The choice probability follows from the standard softmax over the utilities of the alternatives in the choice set:

$$P_k\left(y_n = j \mid \boldsymbol{X}_n, \boldsymbol{s}_{n,k}, \boldsymbol{g}_n\right) = \frac{exp\left(V_{n,j,k}\right)}{\sum_{j' \in \mathcal{J}} exp\left(V_{n,j',k}\right)} \tag{11}$$

The choice loss is the negative log-likelihood:

$$\mathcal{L}_{choice,k} = -\frac{1}{N}\sum_{n=1}^{N} log\, P_k\left(y_n \mid \boldsymbol{X}_n, \boldsymbol{s}_{n,k}, \boldsymbol{g}_n\right) \tag{12}$$

Through the structure-weighted interaction $\boldsymbol{s}_{n,k}$, gradients from $\mathcal{L}_{choice,k}$ propagate to $\boldsymbol{G}_k$ and $\boldsymbol{B}_k$, allowing choice prediction to inform structure learning together with the structural likelihood $\mathcal{L}_{struct,k}$.

**4.4 Joint training**

For a single particle $k$, the total training loss combines the choice and structural objectives with the graph prior, sparsity regularization, the acyclicity constraint, and an economic sign penalty discouraging positive LOS marginal utilities, as shown in **Equation 13**.

$$\begin{aligned} \mathcal{L}_k = {} & \mathcal{L}_{choice,k} + \lambda_{struct}\mathcal{L}_{struct,k} + \lambda_Z \mathcal{L}_Z(\boldsymbol{Z}_k) \\ & + \lambda_{sign}\frac{1}{ND_G}\sum_{n=1}^{N}\sum_{\ell=1}^{D_G} ReLU\,\left(\beta_{n,k}^{(\ell)}\right)^2 \\ & + \lambda_{L1}\sum_{i,r} \boldsymbol{G}_k\,[i,r] + \frac{1}{2}\rho(t)h(\boldsymbol{G}_k)^2 + \mu(t)h(\boldsymbol{G}_k). \end{aligned} \tag{13}$$

where $\lambda_{struct}, \lambda_Z, \lambda_{sign} \geq 0$ are fixed weights, with $\lambda_{sign} = 1$. The flexible specification in **Equation 8** can yield positive coefficients for LOS attributes that represent travel burdens; the sign penalty is therefore included to encourage behaviorally plausible non-positive sensitivities. It is applied to the individual-specific LOS coefficients $\beta_{n,k}^{(\ell)}$, penalizing only positive values through the squared ReLU term while leaving non-positive values unaffected. As a soft regularizer, it can pull positive estimates toward zero relative to unconstrained estimation, but does not impose an exact sign restriction.

The graph-embedding prior is $\mathcal{L}_Z(\boldsymbol{Z}_k) = (\| \boldsymbol{U}_k \|^2 + \| \boldsymbol{V}_k \|^2)/(2\sigma_Z^2)$ with $\sigma_Z^2 = 1/d$ (Lorch et al., 2021). Here, $d$ is the latent embedding dimension defined in **Section 4.2**. Scaling the component-wise variance by $1/d$

keeps the expected squared norm of each $d$-dimensional embedding vector constant as $d$ changes, thereby providing a dimension-normalized prior scale. Because flexible graph search can improve structural fit by adding weakly supported edges, the sparsity penalty discourages unnecessary graph complexity by penalizing the expected number of edges, $\sum_{i,r} \boldsymbol{G}_k[i,r]$. Its weight is scaled with the number of structural observations using the BIC form, $\lambda_{L1} = \lambda_{struct} \frac{1}{2} ln\, N_{struct}$, so that additional edges require sufficient improvement in structural fit to offset their complexity penalty.

The final two terms implement an augmented-Lagrangian treatment of acyclicity. Here, $t$ indexes periodic outer updates of the multiplier $\mu(t)$ and penalty coefficient $\rho(t)$, which are held fixed during the intervening parameter-optimization steps. At each outer update, the remaining acyclicity violation is evaluated across particles; if it exceeds a prescribed tolerance, $\mu(t)$ and $\rho(t)$ are increased. This makes any remaining nonzero $h(\boldsymbol{G}_k)$ contribute a progressively larger linear and quadratic penalty to the training loss. Because $h(\boldsymbol{G}_k) = 0$ only for an acyclic graph, the increasing penalty drives the learned soft adjacency toward the acyclicity constraint.

The two data losses ($\mathcal{L}_{choice}$ and $\mathcal{L}_{struct}$) are coupled through the graph and structural-coefficient parameters $(\boldsymbol{Z}_k, \boldsymbol{B}_k)$. The structural loss $\mathcal{L}_{struct,k}$ evaluates these parameters through the mixed-type structural likelihood (**Equations 3-5**), while $\mathcal{L}_{choice,k}$ influences them indirectly through the structure-weighted interaction (**Equation 7**). In implementation, the choice loss and sign penalty are estimated from minibatches of choice observations, whereas the structural loss is evaluated over the full sample of unique attribute records. Joint training is therefore a single-objective optimization in which both losses shape the posterior over the structural parameters, rather than a heuristic combination of two estimation procedures, an approach shown to be effective in recent transportation studies (Kim et al., 2026b; Lee et al., 2026). The structural-loss weight $\lambda_{struct}$ is selected from the candidate grid based on acyclicity and held-out structural log-likelihood. The reported checkpoint is then chosen using validation weighted F1 among admissible epochs. This procedure selects $5 \times 10^{-2}$ on SP and $1 \times 10^{-4}$ on RP; **Appendix C.3** reports the behavior of the selection criteria across the candidate grid.

Let $\boldsymbol{\theta}_k$ collect the trainable parameters of particle $k$, comprising the graph quantities $\boldsymbol{Z}_k = (\boldsymbol{U}_k, \boldsymbol{V}_k)$, the structural coefficients $\boldsymbol{B}_k$, the structural intercepts $\boldsymbol{b}_k$, and the utility parameters introduced in **Section 4.3**. Rather than estimating a single parameter vector, SVGD maintains $K$ interacting particles, each representing one candidate configuration of the structural and choice-model parameters. The particles are jointly transported toward regions favored by the target posterior while being kept sufficiently distinct from one another, allowing the ensemble to approximate posterior uncertainty without conventional sampling. The update for particle $k$ is defined by the Stein velocity field:

$$\boldsymbol{\phi}(\boldsymbol{\theta}_k) = \frac{1}{K} \sum_{k'=1}^{K} \left[ \kappa(\boldsymbol{\theta}_{k'}, \boldsymbol{\theta}_k) \nabla_{\boldsymbol{\theta}_{k'}} log p(\boldsymbol{\theta}_{k'}) + \nabla_{\boldsymbol{\theta}_{k'}} \kappa(\boldsymbol{\theta}_{k'}, \boldsymbol{\theta}_k) \right] \tag{14}$$

In **Equation 14**, $k$ indexes the particle being updated, whereas $k'$ indexes each particle contributing to its update; $K$ is the total number of particles, $\kappa$ is the kernel measuring similarity between particles, and $\nabla_{\boldsymbol{\theta}_{k'}}$ denotes the gradient with respect to the parameters of particle $k'$. The target log posterior is represented, up to an additive constant, by the negative total loss from **Equation 13**: $log\ p(\boldsymbol{\theta}_k) = -\mathcal{L}_k$, so that $\nabla_{\boldsymbol{\theta}_k} logp(\boldsymbol{\theta}_k) = -\nabla_{\boldsymbol{\theta}_k}\mathcal{L}_k$.

The SVGD update combines two complementary components. The first term is an attractive posterior-score term: it moves particle $k$ toward parameter regions with higher target posterior density, with the contribution of particle $k'$ weighted by their kernel similarity. The second is a repulsive kernel term: it separates nearby particles and prevents the ensemble from collapsing to a single solution. Together, these terms allow the particles to concentrate around well-supported parameter configurations while retaining variation across alternative structural and utility specifications.

To measure similarity between particles without allowing one parameter group to dominate the comparison, we use separate Gaussian radial basis function (RBF) kernels for the structural and choice-model parameter blocks, with bandwidths determined by the median heuristic (Liu and Wang, 2016). The structural kernel is evaluated on $(\boldsymbol{U}_k, \boldsymbol{V}_k, \boldsymbol{B}_k, \boldsymbol{b}_k)$, and the choice-model kernel on the remaining utility parameters. The two kernels are combined in **Equation 14** so that particle attraction reflects similarity in both parts of the model, while repulsion is evaluated on the corresponding parameter block. Parameters are updated along the velocity field as $\boldsymbol{\theta}_k \leftarrow \boldsymbol{\theta}_k + \varepsilon \cdot \boldsymbol{\phi}(\boldsymbol{\theta}_k)$ with step size $\varepsilon$. At convergence, the $K$ particles approximate the posterior over $\boldsymbol{\theta}$; posterior summaries, including edge probabilities, mean coefficients, and mean choice probabilities, are computed by averaging across the ensemble.

### 4.5 Policy simulation

A central motivation for learning the directed dependency structure jointly with the choice model is the ability to simulate policy interventions that respect both layers of the system. Let $q$ denote the intervention-target attribute. For an intervention on individual attribute $q$, the framework executes a topological action step under the learned DAG structure (analogous to Pearl's *do*-operator (Pearl, 2009)): the target is set to $X_{n,q,k}^{do} = x^*$, and the influence of its parent nodes is removed. The remaining descendants are then updated sequentially in topological order of the learned DAG. During this propagation, continuous descendants are updated by preserving their baseline structural disturbances as implied by the fitted model, whereas Bernoulli descendants are sampled from the conditional probabilities implied by their intervened parent values. The resulting responses are averaged over repeated simulation draws to summarize the policy response under the estimated structural and choice components.

For policy simulation, each particle's soft graph is binarized at the threshold 0.5, corresponding to the midpoint of the sigmoid edge probability. Let $\boldsymbol{W}_k = \mathbb{1}[\boldsymbol{G}_k > 0.5] \odot \boldsymbol{G}_k \odot \boldsymbol{B}_k$ denote the structural coefficient matrix used during propagation. The $\boldsymbol{G}_k$ factor is retained intentionally so that thresholding selects the active edges while preserving their learned soft weights. The production graphs at this threshold are acyclic, allowing descendants to be evaluated in a valid topological order. The structural predictor has the same linear form as **Equation 3**, but is evaluated using $\boldsymbol{W}_k$ during policy simulation. Let $\eta_{n,i,k}^{\text{obs}}$ denote this predictor evaluated at the observed parent values and $\eta_{n,i,k}^{\text{do}}$ the predictor evaluated at the currently propagated parent values. For a Gaussian descendant, the fitted baseline residual, $X_{n,i} - \eta^{\text{obs}}_{n,i,k}$, represents the observation-specific realization of the unobserved structural disturbance. Holding this residual fixed while updating the conditional mean yields the following mixed-type update for $i \neq q$:

$$\begin{aligned} X^{\text{do}}_{n,i,k} &= X_{n,i} + \eta^{\text{do}}_{n,i,k} - \eta^{\text{obs}}_{n,i,k}, && i \text{ is Gaussian}, \\ X^{\text{do}}_{n,i,k} &\sim \text{Bernoulli}\left(\sigma\left(\eta^{\text{do}}_{n,i,k}\right)\right), && i \text{ is Bernoulli}. \end{aligned} \tag{15}$$

**Equation 15** preserves this fitted baseline residual for Gaussian descendants, retaining the observation-specific deviation from the fitted structural mean, while Bernoulli descendants are generated as binary realizations from their post-intervention conditional probabilities. The full topological propagation is repeated over multiple Monte Carlo draws to account for the stochastic Bernoulli updates.

After propagation, the structure-weighted interaction is recomputed from the post-intervention attributes:

$$\boldsymbol{s}^{\text{do}}_{n,k} = \boldsymbol{X}^{\text{do}}_{n,k} \odot \left(\boldsymbol{A}_{std,k} \boldsymbol{X}^{\text{do}}_{n,k}\right) \tag{16}$$

**Equations 8-11** are then re-evaluated for each simulation draw by replacing $\boldsymbol{X}_n$ and $\boldsymbol{s}_{n,k}$ with $\boldsymbol{X}^{\text{do}}_{n,k}$ and $\boldsymbol{s}^{\text{do}}_{n,k}$, respectively. Consequently, the propagated attributes enter all components of the LOS coefficients and ASCs through the same utility specification used during estimation.

For compactness, let $P_{n,j,k} \equiv P_k\left(y_n = j \mid \boldsymbol{X}_n, \boldsymbol{s}_{n,k}, \boldsymbol{g}_n\right)$ and $P^{\text{do}}_{n,j,k} \equiv P_k\left(y_n = j \mid \boldsymbol{X}^{\text{do}}_{n,k}, \boldsymbol{s}^{\text{do}}_{n,k}, \boldsymbol{g}_n\right)$ denote the baseline and post-intervention choice probabilities, respectively. The Monte Carlo-averaged change in the probability of alternative $j$ is:

$$\Delta P_{n,j,k} = \mathbb{E}_{\text{MC}}\left[P^{\text{do}}_{n,j,k}\right] - P_{n,j,k} \tag{17}$$

The corresponding downstream attribute responses are computed analogously as $\mathbb{E}_{\text{MC}}\left[X^{\text{do}}_{n,i,k}\right] - X_{n,i}$. Let $N_{\text{int}}$ denote the number of observations over which the intervention is evaluated. For each particle $k$, the scenario-level mode-share and downstream-attribute responses are:

$$\Delta \bar{P}_{j,k} = \frac{1}{N_{\text{int}}} \sum_{n=1}^{N_{\text{int}}} \Delta P_{n,j,k}, \qquad \Delta \bar{X}_{i,k} = \frac{1}{N_{\text{int}}} \sum_{n=1}^{N_{\text{int}}} \left(\mathbb{E}_{\text{MC}}\left[X^{\text{do}}_{n,i,k}\right] - X_{n,i}\right) \tag{18}$$

The final ensemble responses are obtained by averaging across the $K$ particles:

$$\Delta \bar{P}_j = \frac{1}{K} \sum_{k=1}^{K} \Delta \bar{P}_{j,k}, \qquad \Delta \bar{X}_i = \frac{1}{K} \sum_{k=1}^{K} \Delta \bar{X}_{i,k} \tag{19}$$

The first quantity is the predicted scenario-level change in mode share for alternative $j$, and the second is the corresponding average downstream response of attribute $i$. Non-descendants retain their observed values, apart from numerical precision. Reported uncertainty bands are the empirical 5th and 95th percentiles of the particle-specific scenario responses, summarizing across-particle variation within the fitted Neural-BSL specification.

## 5. Results

### 5.1 Predictive performance

We compare the proposed model against two benchmarks on each dataset: a specification-matched MNL model and a feedforward neural network (multilayer perceptron, MLP). The MNL benchmark uses the same individual-attribute vector and the same pairwise interaction set as the proposed model and serves as the warm-start estimator for the linear utility components of Neural-BSL. Specification details, hyperparameter selection, and full coefficient estimate for both benchmarks are reported in **Appendix C**.

**Figure 3** compares how the three models transform the same underlying individual and LOS information into systematic utility. In the MNL, individual attributes enter through prespecified linear and pairwise terms, while LOS attributes enter the utility directly. Neural-BSL preserves this explicit random-utility construction and the direct LOS pathway, but augments the processing of individual attributes with structure-weighted and nonlinear effects that shape the LOS coefficients and ASCs. In contrast, the MLP concatenates the individual and LOS attributes and maps them jointly to systematic utilities through a flexible feedforward network. The comparison therefore holds the underlying predictive information fixed while varying how that information is represented and transformed into utility.

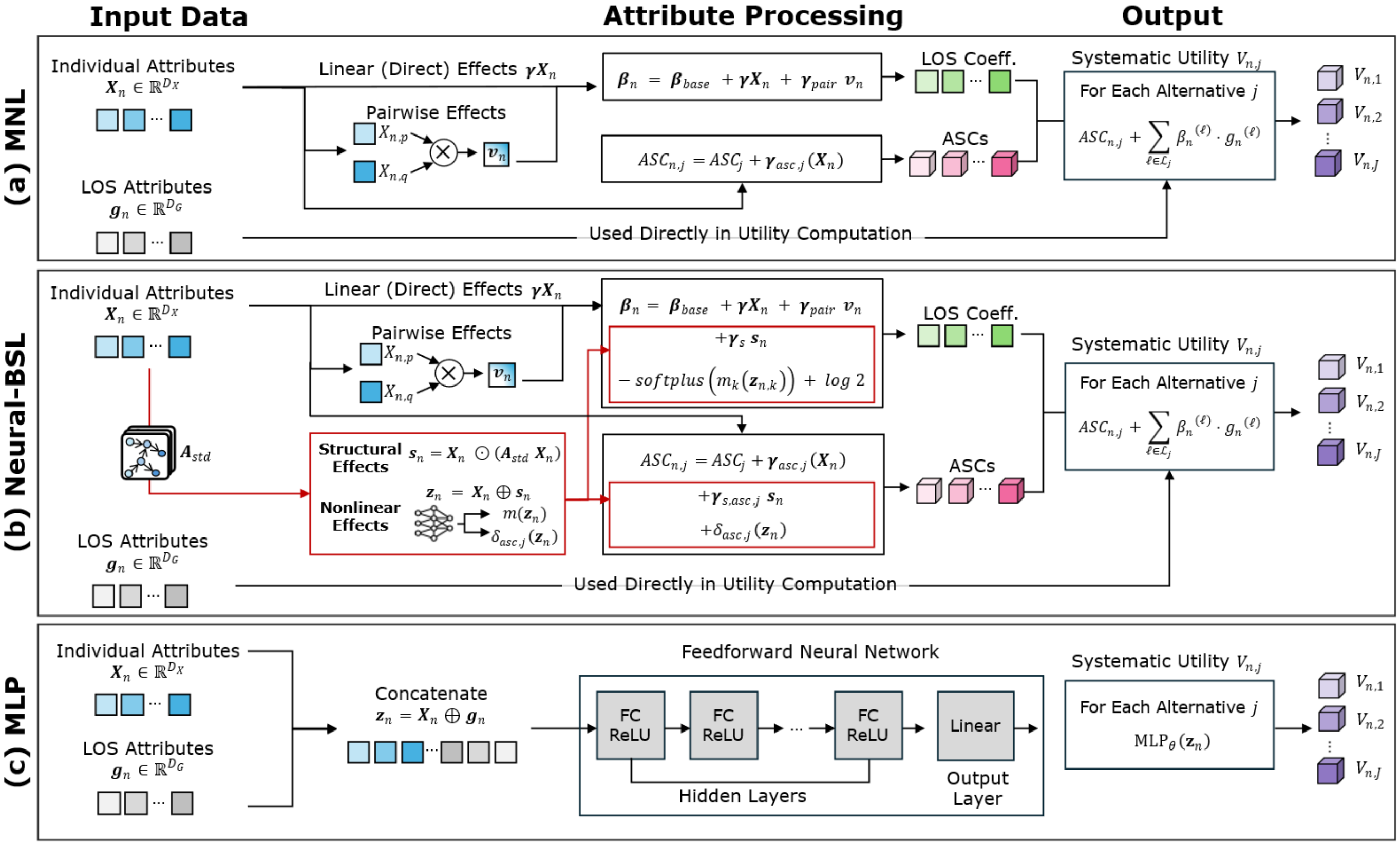

**Figure 3**. Comparison of systematic utility construction from common individual and LOS attributes: (a) MNL, (b) Neural-BSL, and (c) MLP.

**Table 3**. Out-of-sample prediction performance under five-fold cross-validation (mean with standard deviation across folds in parentheses).

| Model | Accuracy | Weighted F1 | Mean log-likelihood |
|---|---|---|---|
| ***SP dataset*** | | | |
| MNL | 0.583 (0.015) | 0.547 (0.015) | -0.916 (0.030) |
| MLP | 0.593 (0.017) | 0.555 (0.020) | -0.898 (0.026) |
| Neural-BSL (Ours) | 0.584 (0.015) | 0.551 (0.017) | -0.916 (0.032) |
| ***RP dataset*** | | | |
| MNL | 0.740 (0.005) | 0.727 (0.006) | -0.681 (0.012) |
| MLP | 0.746 (0.007) | 0.733 (0.008) | -0.659 (0.010) |
| Neural-BSL (Ours) | 0.742 (0.006) | 0.729 (0.006) | -0.675 (0.010) |

**Table 3** reports out-of-sample accuracy, weighted F1, and the average log-likelihood per choice observation. The three models produce mean accuracies within a narrow band on each dataset. On the SP dataset, the spread between the lowest (MNL, 0.583) and highest (MLP, 0.593) is approximately 1.0 percentage point, which is small relative to the cross-fold variation. On the RP dataset, the spread between MNL (0.740) and MLP (0.746) is 0.6 percentage points, with Neural-BSL placing in between at 0.742. The MLP attains the highest accuracy on both datasets, consistent with its unconstrained functional form placing no parametric restriction on how attributes enter the utility.

A correctly specified mode-choice model should assign non-positive marginal utility to LOS attributes that impose traveler costs: longer in-vehicle or access time, higher fares, and additional transfers should not increase an alternative's utility. Because Neural-BSL replaces part of the linear utility with a learned nonlinearity, this sign guarantee is no longer structurally inherent and has to be checked on the fitted model to confirm that the sign penalty in **Equation 13** successfully regularized the parameter space. **Figure 4** reports the alternative-specific own-effect signs recovered by Neural-BSL for the eleven level-of-service attributes in the SP utility and the eleven in the RP utility, each measured as the change in the own-mode choice probability under a one-standard-deviation increase in that attribute. All eleven Neural-BSL own-effects on SP and all eleven on RP carry the theoretically expected non-positive sign, and the mean utility coefficients agree in sign in every case.

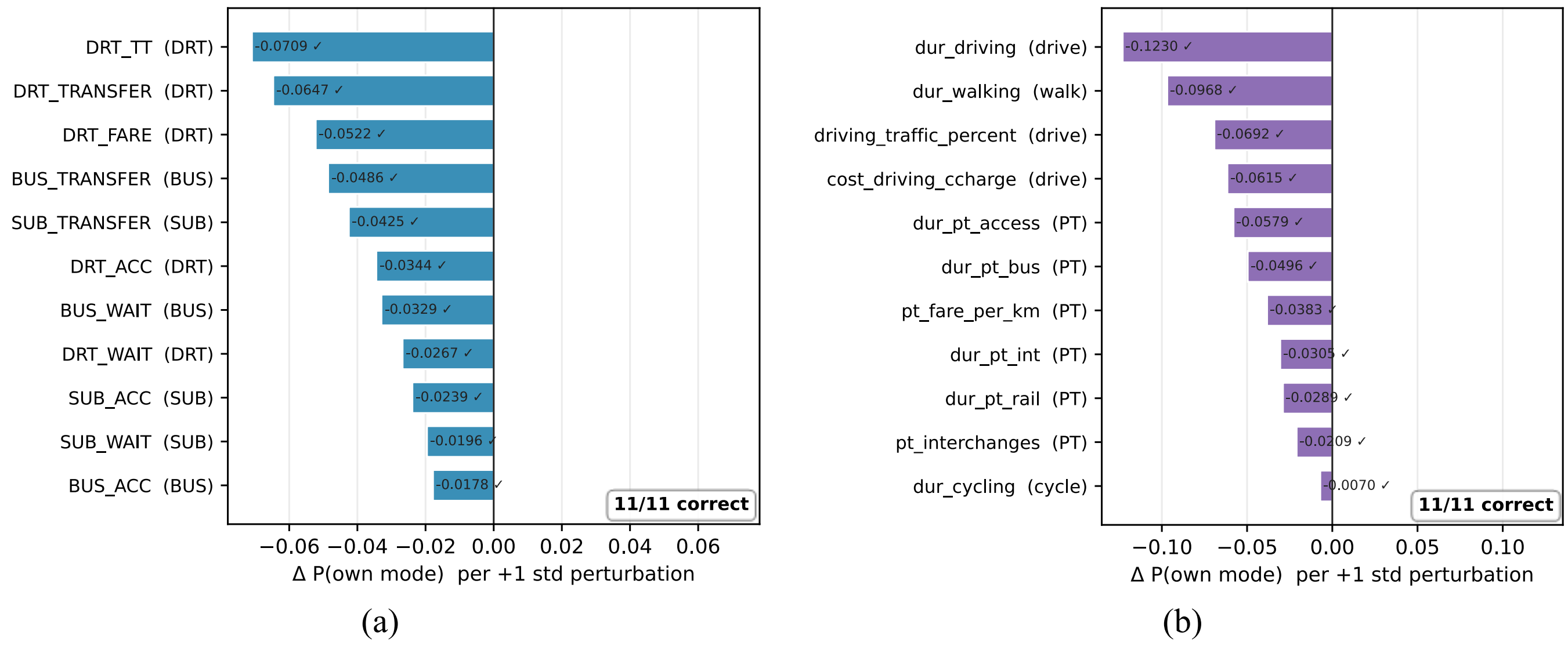

**Figure 4**. Own-effect signs of the LOS attributes under Neural-BSL: (a) SP, (b) RP.

The three models differ less in headline accuracy than in what each model is able to reveal about the choice-generating process. The MNL benchmark provides behaviorally interpretable utility coefficients, from which probability responses and marginal effects can be derived, but it imposes a linear-in-parameters utility and cannot represent nonlinear preference responses or structural dependencies among traveler attributes. The MLP can accommodate flexible nonlinear relationships, but it does not decompose choice probabilities into behaviorally interpretable utility components, fail to impose theoretically consistent signs on LOS effects, and cannot represent the directed dependencies among individual covariates necessary for counterfactual analysis. Neural-BSL bridges these paradigms. It retains an explicit random-utility specification and interpretable linear utility components, while a learned residual captures additional nonlinear variation. Furthermore, the individual covariates are no longer treated as a flat input vector but as nodes in a directed dependency structure estimated jointly with the choice component, enabling structurally grounded counterfactual policy simulations.

### 5.2 Discovered dependency structure

Discrete choice models typically treat individual covariates as an exogenous, flat input vector to the utility function: their joint distribution in the sampled population is taken as given, and the modeler's task is to specify how each covariate enters the utility of each alternative. The covariates themselves, however, are often related to one another (i.e., exhibit rich structural dependencies). For example, driver’s license possession may depend on underlying demographic characteristics, while household car ownership may subsequently depend on driver’s license possession and income. The structural equation tradition in travel demand modeling addresses such relationships by specifying a pre-defined recursive ordering of the covariates a priori and estimating the implied path coefficients (Ben-Akiva and Lerman, 1985; Golob, 2003). Neural-BSL replaces this hand-specified ordering with a directed dependency structure learned directly from data and optimized jointly alongside the utility coefficients.

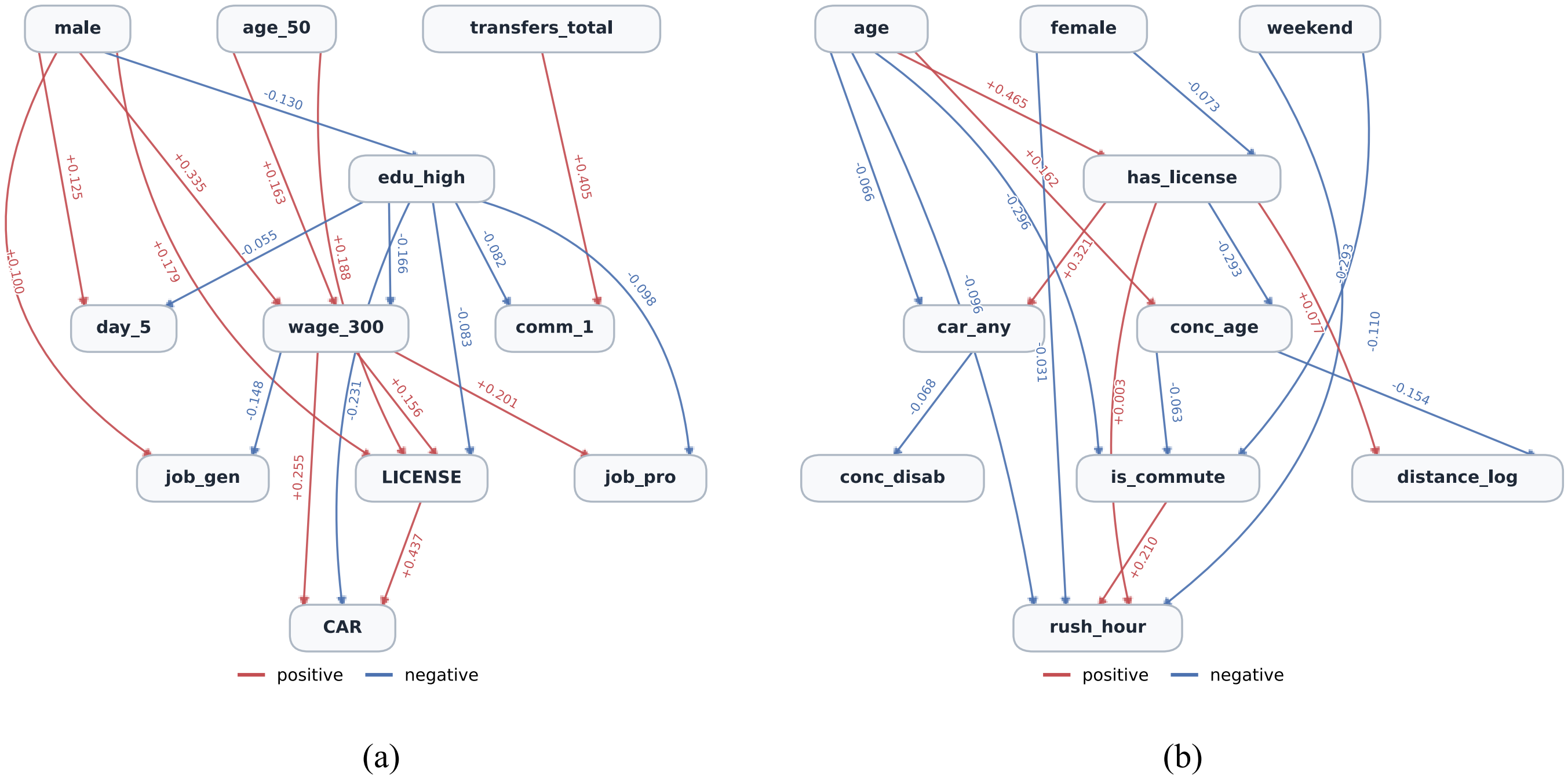

(a) (b)

**Figure 5**. Posterior-summary DAGs for the (a) SP and (b) RP datasets.

**Table 4**. Robust edges discovered in the SP and RP posteriors

| **Parent** | **Child** | $P_{edge}$ | **K-agree** | $A_{std}$ |
|---|---|---|---|---|
| ***Panel A. SP dataset*** | | | | |
| LICENSE | CAR | >0.999 | 10/10 | +0.437 |
| transfers_total | comm_1 | 0.651 | 10/10 | +0.405 |
| male | wage_300 | 0.997 | 10/10 | +0.335 |
| wage_300 | CAR | 0.972 | 10/10 | +0.255 |
| edu_high | CAR | >0.999 | 10/10 | −0.231 |
| wait_15min | walk_15min | 0.566 | 10/10 | +0.203 |
| wage_300 | job_pro | 0.933 | 10/10 | +0.201 |
| age_50 | LICENSE | 0.999 | 10/10 | +0.188 |
| male | LICENSE | 0.932 | 10/10 | +0.179 |
| edu_high | wage_300 | 0.903 | 10/10 | −0.166 |
| age_50 | wage_300 | 0.888 | 10/10 | +0.163 |
| wage_300 | LICENSE | 0.854 | 10/10 | +0.156 |
| wage_300 | job_gen | 0.941 | 10/10 | −0.148 |
| male | edu_high | 0.681 | 10/10 | −0.130 |
| male | day_5 | 0.687 | 10/10 | +0.125 |
| male | job_gen | 0.905 | 9/10 | +0.100 |
| edu_high | job_pro | 0.775 | 10/10 | −0.098 |
| edu_high | LICENSE | 0.911 | 10/10 | −0.083 |
| edu_high | comm_1 | 0.584 | 9/10 | −0.082 |
| edu_high | day_5 | 0.691 | 10/10 | −0.055 |
| ***Panel B. RP dataset*** | | | | |
| age | has_license | >0.999 | 10/10 | +0.465 |
| has_license | car_any | >0.999 | 10/10 | +0.321 |
| age | is_commute | >0.999 | 10/10 | −0.296 |
| weekend | is_commute | >0.999 | 10/10 | −0.293 |
| has_license | conc_age | 0.800 | 8/10 | −0.293 |

| is_commute | rush_hour | >0.999 | 10/10 | +0.210 |
|---|---|---|---|---|
| age | conc_age | 0.807 | 8/10 | +0.162 |
| conc_age | distance_log | 0.874 | 10/10 | −0.154 |
| weekend | rush_hour | 0.989 | 10/10 | −0.110 |
| age | rush_hour | 0.823 | 10/10 | −0.096 |
| has_license | distance_log | 0.713 | 8/10 | +0.077 |
| female | has_license | 0.797 | 8/10 | −0.073 |
| car_any | conc_disab | 0.782 | 8/10 | −0.068 |
| age | car_any | 0.727 | 8/10 | −0.066 |
| conc_age | is_commute | 0.873 | 9/10 | −0.063 |
| female | rush_hour | 0.788 | 8/10 | −0.031 |
| has_license | rush_hour | 0.799 | 8/10 | +0.003 |

* Notes: $A_{std}$ is the change in the child, in standard deviations, per standard deviation of the parent. The edge probability $\boldsymbol{P_{edge}}$ is computed via a sigmoid function and is strictly less than one; entries reported as >0.999 exceed single-precision resolution.

The Bayesian structure learning module (**Section 4.2**) represents posterior uncertainty through $K = 10$ variational particles rather than selecting a single graph. **Figure 5** presents the posterior-summary DAGs for the SP and RP datasets, while **Table 4** reports the full set of robust directed edges. An edge is retained in the posterior-summary when its edge probability exceeds 0.5 (i.e., majority posterior support) and at least eight of the ten particles retain the same directed edge. The three quantities in **Table 4** describe distinct structural properties of each edge. The posterior edge probability $(P_{edge})$ quantifies overall posterior confidence in the directed edge connection, while K-agree records how consistently that edge appears across the ten particles, measuring topological stability across the variational particles. The standardized structural weight $A_{std}$ rescales mixed-type generalized linear score parameters to a common scale (**Equation 6**), so that edges involving differently measured variables can be compared. A positive value indicates a positive relationship from the parent to the child in the learned structure, whereas a negative value indicates an opposite relationship; the absolute value reflects the relative strength of that edge on the standardized scale.

For example, the SP edge from driver's license possession (LICENSE) to car ownership (CAR) has $P_{edge} >$ 0.999, K-agree = 10/10, and $A_{std}$ = 0.437. The first two values indicate that the LICENSE-to-CAR direction receives very strong support and is recovered by every particle. The value of 0.437 further indicates a relatively strong positive relationship on the standardized structural scale. In practical terms, the learned structure indicates a positive relationship between license possession and car ownership. The common standardized scale also allows this value to be compared with, for example, the transfers_total-to-comm_1 edge of 0.405. Edges omitted from **Table 4** may still occur in individual particles but do not meet the robustness criterion used for the posterior summary.

In the SP graph, the most prominent relationships form a mobility-resource pathway connecting socioeconomic characteristics, driver’s license possession, and car ownership. Monthly income at or above KRW 3 million (wage_300) has positive edges to car ownership ($A_{std}$ = 0.255), professional employment (job_pro; 0.201), and license possession (0.156). Age 50 or above (age_50) also has positive edges to income (0.163) and license possession (0.188), whereas high-school education or below (edu_high) has negative edges to income (−0.166) and car ownership (−0.231). The male indicator (male) has positive edges to income (0.335) and license possession (0.179). Together with the strong LICENSE-to-CAR edge described above, these discovered relationships embed private vehicle availability within an upstream socioeconomic hierarchy rather than treating it as an isolated traveler characteristic.

The strongest relationship involving the respondent's typical commuting conditions runs from total transfers in the typical commute (transfers_total) to commuting for at least 60 minutes (comm_1), with $A_{std} = 0.405$. These variables describe the respondent's habitual commuting conditions and are distinct from the alternative-specific LOS attributes presented in the SP choice tasks. The relationship is therefore consistent with greater habitual transfer exposure being associated with a higher likelihood of commuting for at least 60 minutes.

The RP graph reveals a similar separation between mobility resources and trip context, although the pathways extend through a different set of variables. Traveler age (age) has a positive edge to driver's license possession (has_license; $A_{std} = 0.465$), which in turn has a positive edge to the presence of any car in the household (car_any; 0.321). Age also has a smaller negative direct edge to car availability (−0.066). The learned relationship between age and vehicle access therefore combines a positive indirect pathway mediated by driver’s license possession with a weaker offsetting direct relationship. Negative edges run from age and weekend travel (weekend) to work- or school-related travel (is_commute; −0.296 and −0.293), while a positive edge runs from commute purpose to rush-hour travel (rush_hour; 0.210). Taken together, these pathways connect demographic and calendar characteristics with both mobility resources and the purpose and timing of travel.

The concession-related part of the RP graph requires somewhat more cautious interpretation because the age-based fare-concession indicator (conc_age) combines several eligibility groups, including younger travelers. It has negative edges to log trip distance (distance_log; −0.154) and work- or education-related travel (−0.063). The negative edge from driver’s license possession to concession status (−0.293) is less directly connected to an institutional mechanism, since driver’s license possession does not determine fare eligibility. We therefore interpret this relationship not as an institutional causal rule, but as the conditional dependency orientation selected by the learned graph among closely related age, licensing, and concession characteristics.

Overall, the recovered graphs show that several traveler attributes that enter the choice model separately are connected through recurring structural pathways. **Figure 5** and **Table 4** establish the form and relative strength of these relationships, but the presence of an edge alone does not indicate how much it contributes to mode choice. **Section 5.3** therefore examines how much of each attribute's fitted contribution is carried through its downstream structural pathways.

### 5.3 Direct and downstream contributions of individual attributes

The edges identified in **Section 5.2** show which attributes are connected, but they do not indicate how strongly those connections contribute to the choice model. An upstream attribute can enter the modeled choice response through two distinct channels. The first is its direct contribution through its own linear terms in the utility specification ($\boldsymbol{X}_n$). The second is a downstream contribution through other attributes connected to it via the structure-weighted interaction ($\boldsymbol{s}_{n,k}$) in the learned dependency structure. For example, driver's license possession contributes directly to the choice model, while its edge to car ownership provides an additional downstream channel modulating vehicle availability.

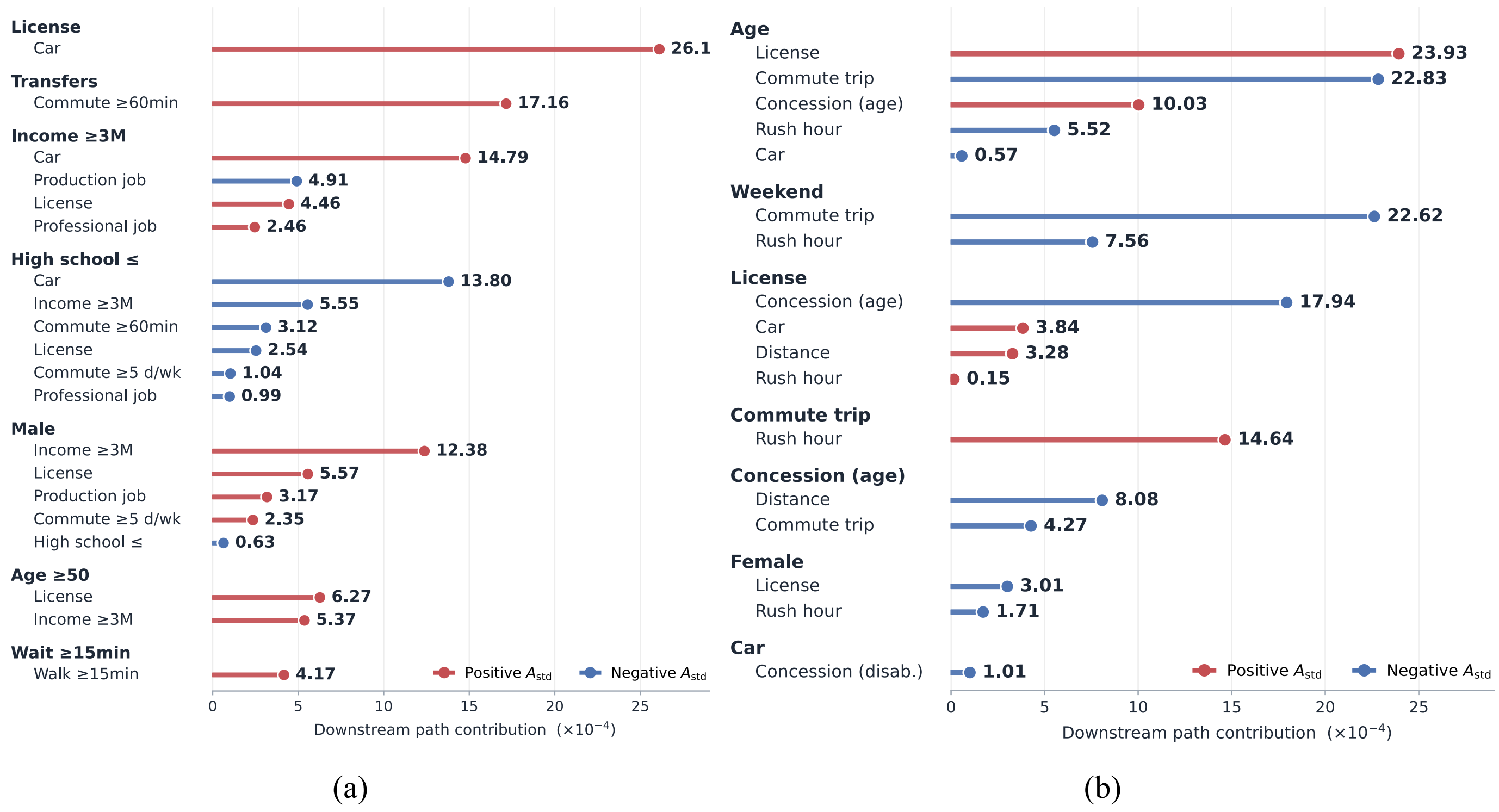

(a) (b)

**Figure 6**. Downstream contributions of robust directed edges to the choice model: (a) SP and (b) RP datasets.

**Figure 6** reports the downstream contribution associated with each robust directed edge. For each edge, this quantity combines its posterior support, standardized structural weight ($\boldsymbol{A}_{\mathrm{std},k}$), and the contribution of the downstream attribute to the structure-weighted choice component in **Equations 8** and **9**. To summarize these relationships at the source-attribute level, the downstream contributions are summed over all outgoing edges from each source and compared with that attribute's direct contribution. The resulting downstream-to-direct ratio indicates the relative importance of the downstream channel for that attribute. Importantly, this decomposition is an in-sample structural attribution that describes how the fitted Neural-BSL model distributes the systematic utility contribution of an observed attribute between its direct and downstream channels, rather than the response produced by changing that attribute in a policy or counterfactual scenario.

In the SP dataset, driver's license possession has the largest downstream-to-direct ratio, at 28.7%, followed by the male indicator at 23.4%, monthly income at or above KRW 3 million at 21.1%, and high-school education or below at 17.9%. The downstream contribution of license possession is concentrated in its edge to car ownership, whereas gender, income, and education occupy broader upstream positions connected to several socioeconomic and mobility-resource attributes. This pattern is behaviorally relevant to the SP setting: although the experimental choice set contains only bus, subway, and DRT, baseline private-vehicle access significantly shapes travelers' latent transit preference and sensitivity. Transit preferences in the SP experiment are therefore related not only to the attributes of the offered modes but also to the broader mobility resources available to the respondent. Total transfers in the typical commute has a downstream-to-direct ratio of 11.4%, with its contribution carried through long commute duration. This provides a separate commute-context channel in which habitual transfer exposure is associated with the broader burden of a long commute, which also contributes to the modeled choice response.

In the RP dataset, traveler age has the largest downstream-to-direct ratio, at 13.9%, followed by weekend travel at 8.5% and driver's license possession at 4.5%. The downstream contribution associated with age extends across both major parts of the RP structure identified in **Section 5.2**: the mobility-resource relationships involving license possession and car availability, and the trip-context relationships involving commute purpose and rush-hour travel. Age therefore enters the fitted RP choice model not only as a traveler characteristic in its own right, but also through differences in mobility resources and in the types and timing of trips associated with travelers of different ages. Weekend travel contributes mainly through the trip-context relationships, indicating that its downstream contribution is associated more with trip purpose and timing than with mobility-resource availability. Work- or education-related travel and age-based fare-concession status have smaller downstream-to-direct ratios of 3.1% and 2.4%, respectively, consistent with their more limited downstream reach.

Across both datasets, the direct contribution remains larger than the summed downstream contribution for every source attribute. The main implication is therefore not that downstream structure replaces the conventional direct interpretation of traveler attributes, but that it adds a secondary channel that can be behaviorally meaningful for attributes such as license possession, age, and commute context. Because each ratio is normalized by the source attribute's direct contribution, it describes the relative balance between the two channels for that attribute and should not be used to rank absolute contributions across attributes or datasets. The ratios characterize the model under the observed data rather than the mode-share response to an exogenous policy intervention. **Section 5.4** therefore examines the selected policy and counterfactual scenarios, where the target change, eligible population, and resulting downstream responses are evaluated together under the *do*-calculus framework.

### 5.4 Mode-share responses under policy and counterfactual scenarios

We evaluate four scenarios for each dataset, combining policy-relevant changes with counterfactuals that can be represented by the variables available in each setting. Ageing and later-life mobility provide the motivation for the age and license scenarios. Seoul is approaching a super-aged population structure, while London is projected to experience substantial growth in its population aged 65 and over (Seoul Metropolitan Government, 2025; Greater London Authority, 2024). Changes in driver's license holding are particularly relevant in this context, given their implications for mobility options in later life (Siren and Haustein, 2015). The remaining scenarios examine habitual commuting conditions, trip purpose, and direct changes in transit fares.

**Figure 7** compares the resulting changes in predicted mode share between Neural-BSL and the specification-matched MNL. Methodologically, when simulating interventions on traveler or trip attributes represented in the DAG, the MNL strictly imposes the *ceteris paribus* assumption by changing only the target attribute while holding all remaining covariates artificially fixed. In contrast, Neural-BSL executes a Pearl-style *do*-intervention (**Section 4.4**): it carries the exogenous change through the learned structural relationships, allowing downstream attributes to adjust along the DAG topology before mode choice probabilities are recomputed. The fare scenarios provide a complementary comparison in which the intervention acts directly through the choice component rather than through the DAG. Changes are reported in percentage points, and the Neural-BSL error bars indicate the empirical 5th–95th percentiles across the ten variational particles.

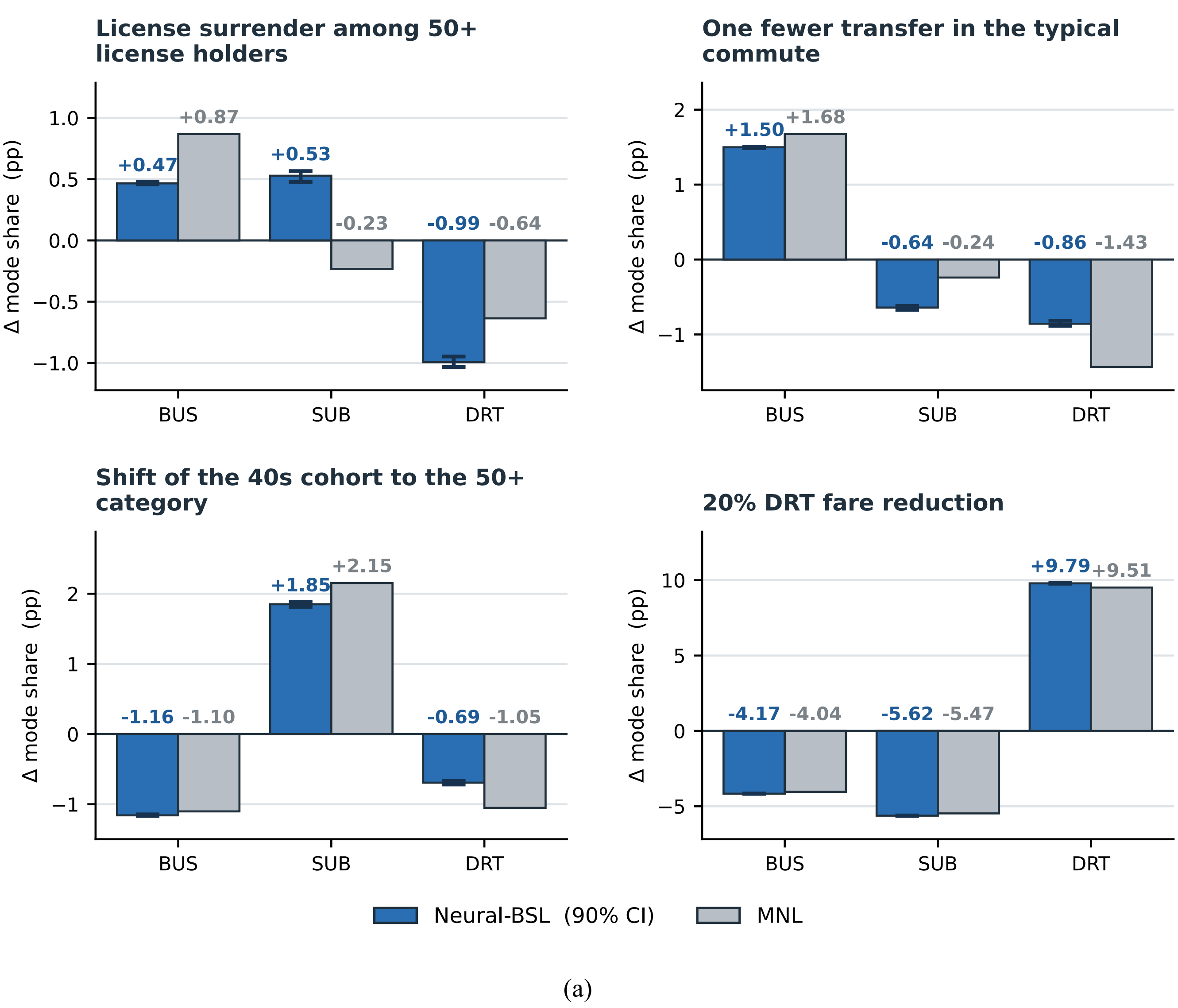

(a)

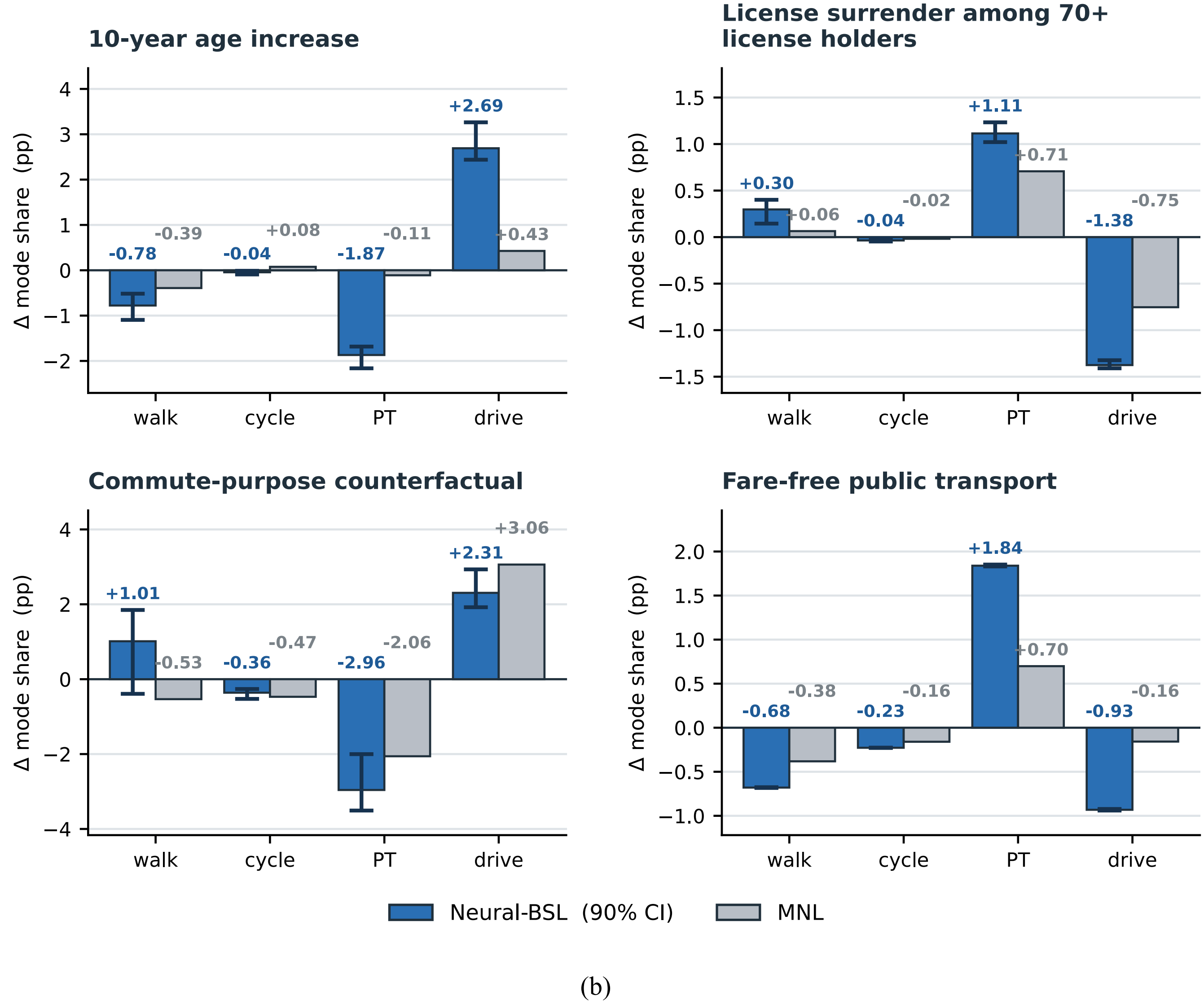

(b)

**Figure 7**. Predicted changes in mode share under the selected policy and counterfactual scenarios: (a) SP and (b) RP

The SP results (**Figure 7(a)**) first illustrate how changes in mobility resources can alter substitution among transit modes. License surrender among license holders aged 50 or above reduces DRT share by 0.99 percentage points (pp) under Neural-BSL, compared with 0.64 pp under the MNL. Neural-BSL reallocates this displaced demand toward both bus (+0.47 pp) and subway (+0.53 pp), whereas the MNL predicts a larger bus increase (+0.87 pp) and a small subway decrease (−0.23 pp). This sign reversal in subway response stems directly from the learned SP DAG, where license possession has a strong edge to car ownership. While the static MNL treats license loss in isolation, Neural-BSL captures the cascading loss of private vehicle availability, realistically shifting car-deprived older travelers toward high-capacity, reliable rail transit (subway) alongside bus.

One fewer transfer in the typical commute represents a change in habitual commuting conditions. The alternative-specific transfer attributes describe the transfers required by the BUS, SUB, and DRT options in the SP task, while the respondent-level transfer measure reflects the transfer environment experienced in the respondent's usual commute. Travelers facing the same modal alternatives may therefore approach the choice task with different commuting backgrounds shaped by their usual transfer exposure and experience. Such differences are behaviorally relevant in Seoul, where transfer penalties have been shown to vary across travel contexts (Yun et al., 2021). The scenario reduces habitual transfer exposure by one while retaining the LOS of the offered alternatives, representing a sustained service or network change that alters the respondent's usual commuting profile. Neural-BSL predicts a +1.50 pp increase in bus share, close to the +1.68 pp predicted by the MNL, but the redistribution across the other modes differs: subway decreases by 0.64 versus 0.24 pp, while DRT decreases by 0.86 versus 1.43 pp. The SP DAG connects habitual transfer exposure with long commuting conditions, meaning that the same modal alternatives are evaluated after a structural shift in the respondent's broader commute profile. This produces a similar overall bus response but a different substitution pattern across subway and DRT.

The SP age counterfactual changes the 50+ age indicator from zero to one for respondents currently in their 40s. Neural-BSL predicts a 1.16 pp decreases in bus and a 1.85 pp increase in subway, compared with −1.10 and +2.15 pp under the MNL. The larger difference appears for DRT, whose predicted reduction is attenuated from 1.05 pp under the MNL to 0.69 pp under Neural-BSL. In the SP DAG, the 50+ age category is upstream of income and license possession, both embedded in the broader mobility-resource structure. The resulting attenuation illustrates that structural relationships can moderate a mode-choice response rather than simply amplify it.

The 20% DRT fare reduction provides a direct service-pricing comparison. Neural-BSL predicts a 9.79 pp increase in DRT, closely matching the MNL estimate of 9.51 pp, with similarly close reductions in bus and subway. DRT fare lies outside the DAG, so the intervention operates directly through the choice component rather than through structural propagation. The same fare change can nevertheless translate into different probability shifts across observations under Neural-BSL, reflecting differences in the traveler profiles entering the choice model. In the SP setting, these differences largely average out, resulting in an aggregate response close to the MNL.

The RP results (**Figure 7(b)**) show stronger differences for several traveler- and trip-context changes. A 10-year age increase raises driving share by 2.69 pp and reduces PT by 1.87 points under Neural-BSL, compared with only +0.43 and −0.11 pp under the MNL. Age occupies an upstream position in the RP DAG, with relationships spanning license possession, car availability, commute purpose, and travel timing. The Neural-BSL response therefore reflects the downstream structural propagation across mobility resources and trip characteristics associated with age in addition to its direct contribution to mode choice, producing a substantially larger redistribution from PT toward driving.

A similar mobility-resource pattern appears under license surrender among license holders aged 70 or above. Neural-BSL predicts a 1.38 pp reduction in driving and a 1.11 pp increase in PT, compared with −0.75 and +0.71 pp under the MNL. The RP DAG contains a strong edge from driver’s license possession to household car availability, placing license status within the broader structure of access to private mobility. Incorporating this relationship produces a stronger shift away from driving than changing the license covariate while retaining the remaining traveler characteristics.

The commute-purpose counterfactual asks how the same traveler might choose a mode if a trip that would otherwise be made for work or education were instead made for another purpose. The observed commute record provides the traveler-specific starting point, while setting the commute-purpose indicator to zero allows the associated trip profile, including travel timing, to adjust through the RP DAG. The simulation can therefore be read as generating a non-commute trip profile for the same traveler rather than simply changing a purpose label. Neural-BSL predicts a 2.96 pp reduction in PT and a 2.31 pp increase in driving, compared with −2.06 and +3.06 pp under the MNL. The broader direction of substitution is similar, but the redistribution differs once the purpose-related trip context is allowed to change.

Fare-free public transport provides the clearest contrast among the direct LOS scenarios. Setting PT fare to zero increases PT share by 1.84 pp under Neural-BSL, compared with 0.70 pp under the MNL, while driving falls by 0.93 rather than 0.16 pp. Approximately 32.5% of RP trips already have zero PT fare, so the intervention primarily affects travelers who currently pay a fare. Since fare lies outside the DAG, this difference does not arise from structural propagation. Instead, the flexible utility specification in Neural-BSL allows fare sensitivity to vary more flexibly across traveler profiles than in the matched MNL specification, translating the same fare removal into a larger aggregate shift toward public transit.

Across all policy scenarios, Neural-BSL diverges meaningfully from the MNL when the intervened attribute is structurally connected to other traveler or trip attributes in the learned DAG. These relationships can alter both the magnitude and the pattern of mode substitution, even reversing response directions as observed in the SP subway outcome. The fare scenarios demonstrate that, even in the absence of DAG propagation, differences between Neural-BSL and the MNL can arise from the more flexible mapping between traveler attributes and LOS sensitivities in the Neural-BSL utility specification.

### 5.5 Downstream covariate responses

The simulations in **Section 5.4** show the mode-share response to each intervention, but they do not by themselves reveal what changes along the learned DAG before the choice probabilities are recomputed. **Figure 8** therefore reports the corresponding responses of downstream traveler and trip attributes, showing which characteristics actually move with each intervention and how these changes relate to the behavioral interpretation of the mode-share results. Values are reported as average standardized changes to allow comparison across differently scaled attributes. Parentheses indicate cases in which the empirical 5th–95th percentile across the ten variational particles includes zero, while dashes indicate no downstream response for the corresponding scenario. The two fare scenarios are omitted because fare lies outside the DAG.

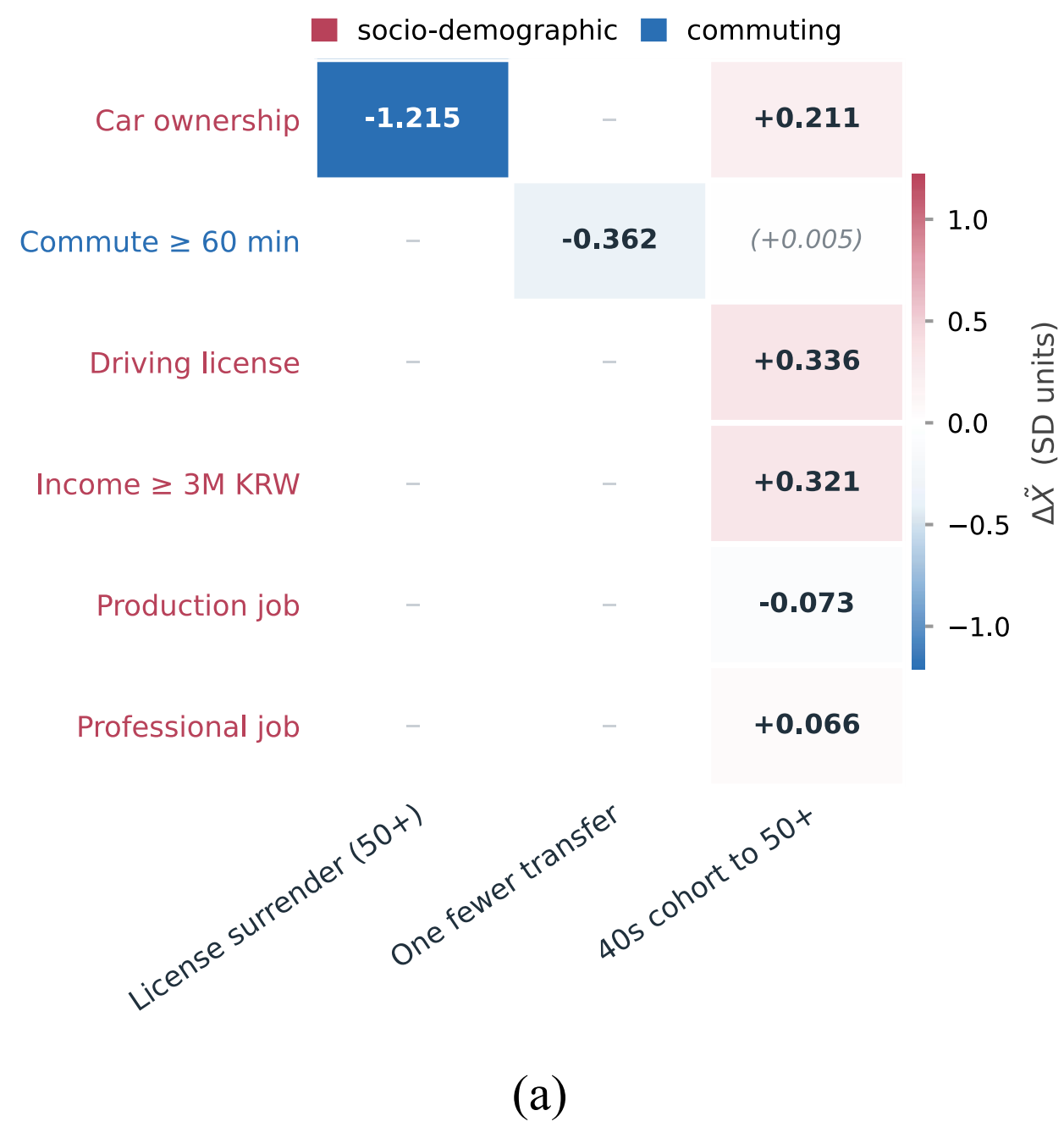

(a)

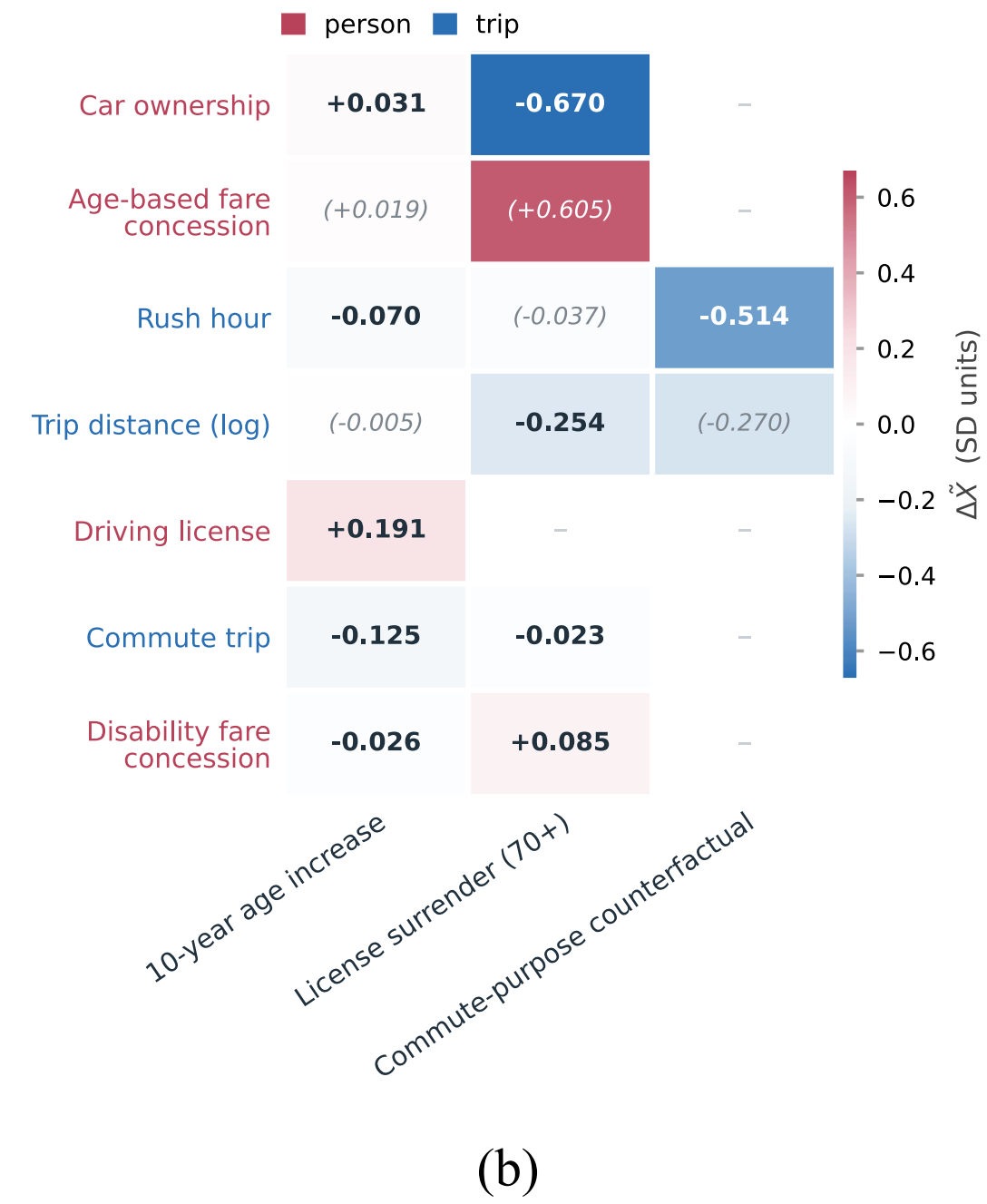

(b)

**Figure 8**. Downstream covariate response: (a) SP, (b) RP.

In the SP dataset (**Figure 8(a)**), license surrender produces the largest downstream response, reducing car ownership by 1.215 standardized units. This is consistent with the strong license-to-car ownership relationship in the discovered structure and shows that changing license status also changes the private-mobility resources associated with the respondent before mode choice is recomputed. This helps explain why Neural-BSL produces not only a larger decline in DRT but also a different subway response from the MNL. One fewer transfer in the typical commute reduces the long-commute indicator by 0.362 standardized units. The intervention therefore changes the respondent's habitual commute conditions even though the alternative-specific LOS remains unchanged, supporting the different redistribution across subway and DRT observed in the mode-share results.

The shift from the 40s cohort to the 50+ category produces increases in license possession (+0.336), higher-income status (+0.321), and car ownership (+0.211). These changes follow the socioeconomic and mobility-resource relationships identified in the SP DAG. The age intervention therefore shifts several characteristics associated with access to private mobility at the same time. These accompanying changes can offset part of the age-related choice response, consistent with the smaller reduction in DRT predicted by Neural-BSL.

In the RP dataset (**Figure 8(b)**), the 10-year age increase affects both mobility resources and trip context. License possession increases by 0.191 standardized units and car availability by 0.031, while work- or education-related travel decreases by 0.125 and rush-hour travel by 0.070. The age change is therefore accompanied by greater access to private mobility as well as a different mix of trip purpose and timing. These joint changes explain why Neural-BSL generates a much more pronounced mode shift from public transit toward driving than the static MNL.

License surrender among older license holders reduces car availability by 0.670 standardized units, providing a clear mechanism for the stronger shift away from driving. Trip distance also decreases by 0.254 standardized units, indicating that the downstream adjustment extends beyond vehicle availability. For the commute-purpose counterfactual, the clearest response is a 0.514-standardized-unit decrease in rush-hour travel. In practical terms, the non-commute trip profile generated by the intervention is substantially less associated with peak-period travel. The average change in trip distance is less stable across particles, making travel timing the more consistent downstream pathway for this scenario.

The downstream responses also explain why DAG propagation does not affect every mode-share prediction in the same way. Changes in connected mobility resources or trip context can reinforce the response to the intervention, offset part of it, or alter how the resulting share is redistributed across alternatives. The discovered DAG therefore affects the simulations through the specific traveler and trip characteristics that adjust after an intervention, rather than through a uniform increase or decrease in the predicted response.

## 6. Conclusions and future work

This paper proposed Neural-BSL, a framework that unifies Bayesian structure learning of traveler attribute dependencies and discrete choice modeling within a single estimation procedure. The directed dependency structure among traveler attributes is represented through differentiable adjacency variables and estimated jointly with the choice model, allowing the graph topology and choice relationships to be learned simultaneously rather than in disconnected stages. The observed choice is excluded from the DAG and connected to the traveler attributes through an augmented random-utility specification. The learned structure also enters the choice model through structure-weighted interactions among the attributes. For policy and counterfactual analysis, exogenous interventions are propagated along the learned DAG via a Pearl-style *do*-operation before the choice probabilities are recomputed. The framework therefore provides both the resulting change in mode share and the accompanying changes in downstream traveler or trip attributes, which are fundamentally missed when a conventional choice simulation changes only the target covariate under static *ceteris paribus* assumptions.

These properties were examined using SP data from the Seoul metropolitan area and RP data from London. First, the dependency structure among traveler attributes is learned from the data rather than fixed by the analyst in advance. Second, the fitted choice model distinguishes an attribute's direct utility contribution from the additional contribution associated with its downstream structural relationships. The empirical results show that these downstream channels capture behaviorally meaningful mediating relationships for attributes such as license possession, age, and commute context. Third, the position of the intervention target in the learned DAG determines which other attributes can adjust before mode choice is recomputed. The simulations show that these downstream adjustments do not affect forecasts in a single direction: depending on the attributes involved, they can strengthen, moderate, or redistribute the predicted mode shift, even reversing substitution directions as observed in the SP subway response. The downstream covariate responses further identify the mobility-resource and trip-context changes underlying these differences.

Neural-BSL has three main limitations. First, its causal interpretation rests on the standard assumption of causal sufficiency over the observed attributes. We do not claim that the framework recovers the true data-generating mechanism in the presence of unmeasured confounders; the discovered DAG should instead be interpreted as the structural conditional-dependency supported by the observed data under the modeling assumptions adopted here. Second, the discovered structure depends on choices made during structural learning, including the regularization setting, the edge-retention criterion, and the prior mask. Neural-BSL represents structural uncertainty through multiple variational particles and carries this variation into the reported empirical particle intervals, but alternative settings may add or remove edges and change the resulting downstream interpretations. Third, the DAG is learned from cross-sectional data, so some edge directions cannot be fully resolved from the data (i.e., observational score functions) alone. The prior mask rules out implausible orientations, but the remaining structure therefore combines empirical evidence with these prior restrictions. Structural learning in the SP application also faces a separate data limitation. Although the dataset contains 5,178 choice observations, these observations come from 863 respondents completing repeated choice tasks, providing substantially fewer distinct respondent profiles for learning relationships among respondent-level attributes. Larger SP samples with more distinct respondent profiles would therefore provide a stronger basis for structural learning, although such data are more demanding to collect.

These limitations suggest several promising directions for future work. Panel or longitudinal data would provide temporal ordering to resolve directional ambiguities that remain in cross-sectional analysis and would allow the downstream responses to be compared with observed changes over time. A second direction concerns the prior-knowledge mask used in structure learning. The mask determines which orientations are admissible before estimation, so its specification can influence both the learned DAG and the interventions derived from it. Large language models have recently been applied to transportation problems involving contextual reasoning and socio-demographic representation, and they may offer a systematic way to help construct or check the validity of the mask, substantially reducing dependence on analyst judgment (Yun and Lee, 2025; Lim et al., 2026). Finally, LOS attributes lie outside the discovered structure in the present model. In the SP data, their values are set by the experimental design rather than by the travelers, so they are excluded from the attribute DAG. In RP settings, however, some LOS attributes (e.g., travel times and costs) may exhibit endogenous structural dependencies with land-use patterns, route choices, or congestion. Extending the framework to represent endogenous LOS attributes where empirically justified would connect the structural responses shown here more directly to the service and pricing levers that mode choice models are commonly used to assess.

## Reference

Andrews, B., Ramsey, J., Cooper, G.F., 2018. Scoring Bayesian networks of mixed variables. Int. J. Data Sci. Anal. 6, 3-18. https://doi.org/10.1007/s41060-017-0085-7

Ben-Akiva, M., Lerman, S.R., 1985. Discrete Choice Analysis: Theory and Application to Travel Demand. MIT Press, Cambridge.

Ben-Akiva, M., Walker, J., Bernardino, A.T., Gopinath, D.A., Morikawa, T., Polydoropoulou, A., 2002. Integration of Choice and Latent Variable Models. Perpetual motion: Travel behaviour research opportunities and application challenges.

Bliemer, M.C.J., Rose, J.M., 2009. Efficiency and Sample Size Requirements for Stated Choice Studies, in: Transportation Research Board 88th Annual Meeting.

Brathwaite, T., Walker, J.L., 2018. Causal inference in travel demand modeling (and the lack thereof). J. Choice Model. 26, 1-18. https://doi.org/10.1016/j.jocm.2017.12.001

Cascetta, E., 2009. Transportation systems analysis: models and applications. Springer Science & Business Media.

Cascetta, E., Papola, A., 2001. Random utility models with implicit availability/perception of choice alternatives for the simulation of travel demand. Transp. Res. Part C Emerg. Technol. 9, 249-263. https://doi.org/10.1016/S0968-090X(00)00036-X

Chauhan, R.S., Riis, C., Adhikari, S., Derrible, S., Zheleva, E., Choudhury, C.F., Marathe, A., 2024. Determining causality in travel mode choice. Travel Behav. Soc. 36, 100789. https://doi.org/10.1016/j.tbs.2024.100789

Chauhan, R.S., Sutradhar, U., Rozhkov, A., Derrible, S., 2025. Causation versus prediction in travel mode choice modeling. npj Sustain. Mobil. Transp. 2, 1-11. https://doi.org/10.1038/s44333-024-00022-4

Chickering, D.M., 2003. Optimal structure identification with greedy search. J. Mach. Learn. Res. 3, 507-554. https://doi.org/10.1162/153244303321897717

El Zarwi, F., Vij, A., Walker, J.L., 2017. A discrete choice framework for modeling and forecasting the adoption and diffusion of new transportation services. Transp. Res. Part C Emerg. Technol. 79, 207-223. https://doi.org/10.1016/j.trc.2017.03.004

Golob, T.F., 2003. Structural equation modeling for travel behavior research. Transp. Res. Part B Methodol. 37, 1-25. https://doi.org/10.1016/S0191-2615(01)00046-7

Greater London Authority, 2024. GLA 2022-based population projections. London Datastore. https://data.london.gov.uk/blog/gla-2022-based-population-projections (accessed 7 August 2026).

Guajardo Ortega, M.F., Link, H., 2025. Mode choice inertia and shock: Three months of almost fare-free public transport in Germany. Econ. Transp. 41. https://doi.org/10.1016/j.ecotra.2024.100382

Hagenauer, J., Helbich, M., 2017. A comparative study of machine learning classifiers for modeling travel mode choice. Expert Syst. Appl. 78, 273-282. https://doi.org/10.1016/j.eswa.2017.01.057

Han, Y., Pereira, F.C., Ben-Akiva, M., Zegras, C., 2022. A neural-embedded discrete choice model: Learning taste representation with strengthened interpretability. Transp. Res. Part B Methodol. 163, 166-186. https://doi.org/10.1016/j.trb.2022.07.001

Hausman, J.A., McFadden, D., 1984. Specification tests for the multinomial logit model. Econometrica 52, 1219-1240. https://doi.org/10.2307/1913827

Hillel, T., Elshafie, M.Z.E.B., Jin, Y., 2018. Recreating passenger mode choice-sets for transport simulation: A case study of London, UK. Proc. Inst. Civ. Eng. Smart Infrastruct. Constr. 171, 29-42. https://doi.org/10.1680/jsmic.17.00018

Huang, B., Zhang, K., Lin, Y., Scholkopf, B., Glymour, C., 2018. Generalized score functions for causal discovery. Proc. ACM SIGKDD Int. Conf. Knowl. Discov. Data Min. 1551-1560. https://doi.org/10.1145/3219819.3220104

Kim, E.-J., Oh, D., Kim, H., 2026a. Causal impacts of demand responsive transit on public transit demand: A spatial assessment framework. Transp. Policy 179, 104008. https://doi.org/10.1016/j.tranpol.2026.104008

Kim, E.J., 2021. Analysis of Travel Mode Choice in Seoul Using an Interpretable Machine Learning Approach. J. Adv. Transp. 2021. https://doi.org/10.1155/2021/6685004

Kim, E.J., Bansal, P., 2024. A new flexible and partially monotonic discrete choice model. Transp. Res. Part B Methodol. 183, 102947. https://doi.org/10.1016/j.trb.2024.102947

Kim, E.J., Kim, Y., Jang, S., Kim, D.K., 2021. Tourists' preference on the combination of travel modes under Mobility-as-a-Service environment. Transp. Res. Part A Policy Pract. 150, 236-255. https://doi.org/10.1016/j.tra.2021.06.016

Kim, J.T., Kim, D.-K., Yun, H., 2026b. Gradient-based calibration of volume delay functions in static traffic assignment via implicit differentiation. SSRN. https://doi.org/10.2139/ssrn.6993254

Kim, Y., Zardini, G., Samaranayake, S., Shafiee, S., 2024. Estimate Then Predict: Convex Formulation for Travel Demand Forecasting. SSRN. https://doi.org/10.2139/ssrn.4977199

Ko, J., Lee, S., Byun, M., 2019. Exploring factors associated with commute mode choice: An application of city-level general social survey data. Transp. Policy 75, 36-46. https://doi.org/10.1016/j.tranpol.2018.12.007

Kuipers, J., Moffa, G., 2025. The interventional Bayesian Gaussian equivalent score for Bayesian causal inference with unknown soft interventions. Proc. Mach. Learn. Res. 275, 772-791.

Lee, D., Derrible, S., Pereira, F.C., 2018. Comparison of Four Types of Artificial Neural Network and a Multinomial Logit Model for Travel Mode Choice Modeling. Transp. Res. Rec. 2672, 101-112. https://doi.org/10.1177/0361198118796971

Lee, E.H., 2022. Exploring Transit Use during COVID-19 Based on XGB and SHAP Using Smart Card Data. J. Adv. Transp. 2022. https://doi.org/10.1155/2022/6458371

Lee, E.H., Kim, K., Kho, S.Y., Kim, D.K., Cho, S.H., 2022. Exploring for route preferences of subway passengers using smart card and train log data. J. Adv. Transp. 2022, 6657486. https://doi.org/10.1155/2022/6657486

Lee, J., Kim, J., 2025. Where can automated mobility-on-demand service thrive: A combined method of latent class choice and random forest. Transp. Policy 165, 127-149. https://doi.org/10.1016/j.tranpol.2025.02.019

Lee, J., Kim, G., Sohn, K., 2026. Transformation of a bi-level transit network design problem into single-objective unconstrained optimization. Netw. Spat. Econ. 26, 189-237.

Lim, S. Y., Yun, H., Bansal, P., Kim, D. K., & Kim, E. J. (2026). A large language model for feasible and diverse population synthesis. Transp. Res. Part C: Emerging Tech. 185, 105581.

Liu, Q., Wang, D., 2016. Stein variational gradient descent: A general purpose Bayesian inference algorithm. Adv. Neural Inf. Process. Syst. 2378-2386.

Lorch, L., Rothfuss, J., Scholkopf, B., Krause, A., 2021. DiBS: Differentiable Bayesian Structure Learning. Adv. Neural Inf. Process. Syst. 34, 24111-24123.

Lundberg, S.M., Lee, S.-I., 2017. A unified approach to interpreting model predictions. Adv. Neural Inf. Process. Syst. 30, 4765-4774.

Ma, T.-Y., Chow, J.Y.J., Xu, J., 2017. Causal structure learning for travel mode choice using structural restrictions and model averaging algorithm. Transportmetrica A: Transp. Sci. 13, 299-325. https://doi.org/10.1080/23249935.2016.1265019

Martin-Baos, J.A., Lopez-Gomez, J.A., Rodriguez-Benitez, L., Hillel, T., Garcia-Rodenas, R., 2023. A prediction and behavioural analysis of machine learning methods for modelling travel mode choice. Transp. Res. Part C Emerg. Technol. 156, 104318. https://doi.org/10.1016/j.trc.2023.104318

McFadden, D., 1974. Conditional logit analysis of qualitative choice behavior, in: Zarembka, P. (Ed.), Frontiers in Econometrics. Academic Press, New York, pp. 105-142.

Natterer, E.S., Rao, S.R., Tejada Lapuerta, A., Engelhardt, R., Horl, S., Bogenberger, K., 2025. Machine learning surrogates for agent-based models in transportation policy analysis. Transp. Res. Part C Emerg. Technol. 180, 105360. https://doi.org/10.1016/j.trc.2025.105360

Park, S.J., Kim, E.-J., 2025. How Do Metropolitan Commuters Respond to Demand Responsive Transit? Evidence from Seoul Metropolitan Area. SSRN. https://doi.org/10.2139/ssrn.5398318

Pearl, J., 2009. Causality. Cambridge University Press.

Robinson, R.W., 1977. Counting unlabeled acyclic digraphs, in: Combinatorial Mathematics V, Lecture Notes in Mathematics 622. Springer, Berlin, pp. 28-43.

Seoul Metropolitan Government, 2025. 9988 Seoul Project for a super-aged society. Seoul Metropolitan Government. https://mediahub.seoul.go.kr/archives/2014598 (accessed 7 August 2026).

Shmueli, G., 2010. To explain or to predict? Stat. Sci. 25, 289-310. https://doi.org/10.1214/10-STS330

Sifringer, B., Lurkin, V., Alahi, A., 2020. Enhancing discrete choice models with representation learning. Transp. Res. Part B Methodol. 140, 236-261. https://doi.org/10.1016/j.trb.2020.08.006

Siren, A., Haustein, S., 2015. Driving licences and medical screening in old age: review of literature and European licensing policies. J. Transp. Health 2(1), 68-78. https://doi.org/10.1016/j.jth.2014.09.003

Spirtes, P., Glymour, C., 1991. An Algorithm for Fast Recovery of Sparse Causal Graphs. Soc. Sci. Comput. Rev. 9, 62-72.

Spirtes, P., Glymour, C., Scheines, R., 2000. Causation, prediction, and search, MIT press.

Train, K.E., 2009. Discrete Choice Methods with Simulation, 2nd ed. Cambridge University Press, Cambridge.

Vij, A., Walker, J.L., 2016. How, when and why integrated choice and latent variable models are latently useful. Transp. Res. Part B Methodol. 90, 192-217. https://doi.org/10.1016/j.trb.2016.04.021

Wang, F., Ross, C.L., 2018. Machine learning travel mode choices: comparing the performance of an extreme gradient boosting model with a multinomial logit model. Transp. Res. Rec. 2672, 35-45. https://doi.org/10.1177/0361198118773556

Wang, S., Mo, B., Zhao, J., 2021. Theory-based residual neural networks: A synergy of discrete choice models and deep neural networks. Transp. Res. Part B Methodol. 146, 333-358. https://doi.org/10.1016/j.trb.2021.03.002

Wong, M., Farooq, B., 2021. ResLogit: A residual neural network logit model for data-driven choice modelling. Transp. Res. Part C Emerg. Technol. 126, 103050. https://doi.org/10.1016/j.trc.2021.103050

Xie, C., Waller, S.T., 2010. Estimation and application of a Bayesian network model for discrete travel choice analysis. Transp. Lett. 2, 125-144. https://doi.org/10.1179/tl.2010.2.2.125

Yu, Y., Chen, J., Gao, T., Yu, M., 2019. DAG-GNN: DAG structure learning with graph neural networks. 36th Int. Conf. Mach. Learn. ICML 2019 2019-June, 12395-12406.

Yun, H., Lee, E. H., Kim, D. K., & Cho, S. H., 2021. Development of estimating methodology for transit accessibility using smart card data. Transp. Res. Rec. 2675, 159-171.

Yun, H., Lee, E.H., Moon, S., Kim, D.K., 2024. Data-Driven Approach for Measuring and Managing Physical Distancing in Subways during Pandemic Conditions. Transp. Res. Rec. 2678, 588-600. https://doi.org/10.1177/03611981231190394

Yun, H., & Lee, E.H. (2025). Party politics in transport policy with a large language model. Transp. Policy. 171, 487-496.

Zheng, X., Aragam, B., Ravikumar, P., Xing, E.P., 2018. Dags with no tears: Continuous optimization for structure learning. Adv. Neural Inf. Process. Syst. 2018-December, 9472-9483.

Zheng, X., Dan, C., Aragam, B., Ravikumar, P., Xing, E.P., 2020. Learning Sparse Nonparametric DAGs. Proc. Mach. Learn. Res. 108, 3414-3425.

## Appendix A. Attitudinal items and factor analysis

The 17 Likert-scale attitudinal items collected in the SP survey are reduced to four continuous factor scores via exploratory factor analysis (EFA) using minimum-residual (minres) extraction with varimax rotation, fitted on the training respondents only (n = 604) to avoid information leakage from the validation and test partitions. The items are grouped a priori into four constructs: punctuality, efficiency, comfort, and innovativeness. The resulting factor scores are used as attitudinal attributes in Neural-BSL in place of the raw items. All 17 items are retained in the factor analysis without ex-ante item exclusion.

Sampling adequacy is satisfactory: the overall Kaiser–Meyer–Olkin (KMO) measure is 0.82, with all per-item KMO values at or above 0.51. Bartlett's test of sphericity rejects the null of an identity correlation matrix ($\chi^2$ = 2,384.27, df = 136, $p < 0.001$), supporting the appropriateness of factor analysis. The four extracted factors cumulatively explain 37.3% of total item variance.

**Table A.1**. Standardized factor loadings for the 17 attitudinal items. Loadings with $|\lambda| \geq 0.4$ are shown in bold. Factor scores are computed using the regression method.

| Code | Item statement | Innovat. | Comfort | Punctual. | Efficiency |
|---|---|---|---|---|---|
| ***Punctuality items*** | | | | | |
| P_1 | I am reluctant to use modes whose travel times vary unpredictably. | −0.094 | −0.155 | **+0.568** | −0.018 |
| P_2 | I prefer reservable modes because they remove waiting uncertainty | +0.219 | +0.293 | +0.163 | +0.183 |
| P_3 | I try to avoid arriving later than planned. | +0.144 | +0.050 | **+0.686** | +0.045 |
| P_4 | When traveling, I tend to allow extra time for unexpected delays. | +0.101 | +0.191 | +0.188 | +0.245 |
| P_5 | I avoid reservable modes because departure times may change | −0.127 | −0.126 | +0.166 | +0.124 |
| ***Efficiency items*** | | | | | |
| E_1 | I prefer modes that minimize travel time. | +0.246 | +0.187 | **+0.420** | +0.324 |
| E_2 | I view waiting time during a trip as wasted time. | +0.095 | +0.081 | **+0.419** | +0.042 |
| E_3 | I am willing to pay more for a faster trip. | +0.367 | **+0.461** | +0.011 | +0.387 |
| E_4 | I am willing to transfer if it shortens my travel time. | +0.183 | −0.016 | +0.042 | **+0.601** |
| ***Comfort items*** | | | | | |
| C_1 | I prefer modes with less in-vehicle crowding. | +0.249 | **+0.555** | +0.166 | +0.066 |
| C_2 | I prefer to travel seated even if it takes longer. | +0.112 | **+0.601** | −0.061 | −0.027 |
| C_3 | I am comfortable traveling with many other passengers. | +0.186 | −0.372 | +0.042 | +0.233 |
| C_4 | I value convenience over cost. | +0.199 | **+0.513** | −0.039 | +0.204 |
| ***Innovativeness items*** | | | | | |
| N_1 | I am open to using new transportation modes. | **+0.829** | +0.068 | +0.034 | +0.089 |
| N_2 | I am comfortable using transport services through mobile apps. | **+0.683** | +0.130 | +0.173 | +0.080 |
| N_3 | I enjoy trying new things. | **+0.601** | +0.196 | −0.004 | +0.303 |
| N_4 | I prefer new modes if they allow faster travel. | **+0.652** | +0.255 | +0.092 | +0.223 |
| *Variance explained (%)* | | *14.30* | *9.51* | *7.72* | *5.80* |
| *Cumulative variance (%)* | | *14.30* | *23.81* | *31.53* | *37.33* |

## Appendix B. Causal plausibility mask

Purely data-driven structure learning can face two practical difficulties. First, observational data alone may not resolve some edge directions because multiple DAGs can belong to the same Markov equivalence class. Second, strong or nearly deterministic associations can favor directions that are difficult to justify from temporal or substantive knowledge. For example, almost every car owner in our data holds a driving license, so predicting license possession from car ownership can fit the data well even though license holding generally precedes vehicle acquisition. We therefore introduce a causal plausibility mask that excludes directions that are difficult to support given temporal ordering, variable construction, or the interpretation adopted in this application. All remaining directions are left for the model to learn from the data.

### SP Dataset mask

Six rules apply to the 17 attributes included in the SP DAG.

**Rule 1:** Gender and age cohort are treated as exogenous and therefore have no parents.

**Rule 2:** Education may have only gender and age cohort as parents, reflecting its earlier position in the life course.

**Rule 3:** Car ownership is not allowed as a parent of license holding, with license acquisition treated as preceding vehicle ownership.

**Rule 4:** The long-commute indicator is not allowed as a parent of the transfer count, because transfer requirements are treated as part of the conditions contributing to commute duration.

**Rule 5:** Car ownership is not allowed as a parent of income, with income treated as the upstream socioeconomic resource in this cross-sectional specification.

**Rule 6:** The two occupation indicators cannot be parents of each other because they are derived from the same categorical occupation variable.

### RP Dataset mask

Five rules apply to the 11 attributes in the RP dataset: 6 person-level attributes and 5 trip-context attributes.

**Rule 1:** The female indicator, age, and the weekend and winter-month indicators are treated as exogenous and therefore have no parents.

**Rule 2:** Trip-context attributes are not allowed as parents of license holding, car availability, or fare-concession status, which are treated as established before the observed trip.

**Rule 3:** Car availability is not allowed as a parent of license holding, with license acquisition treated as preceding access to a household car.

**Rule 4:** Rush-hour travel is not allowed as a parent of commute status, with trip purpose treated as preceding departure timing.

**Rule 5:** The age-based and disability-based fare-concession indicators cannot be parents of each other because they are mutually exclusive categories derived from the same fare-type variable.

## Appendix C. Benchmark specifications and sensitivity analysis

### C.1 Specification-matched MNL

The MNL benchmark matches the linear utility specification used to initialize Neural-BSL: LOS coefficients, ASCs, mode-specific individual-attribute coefficients, and LOS-by-individual interactions, estimated by maximum likelihood with L2-regularized interaction terms. Its fitted coefficients are used to warm-start the corresponding linear utility components of Neural-BSL, so the model begins near the specification-matched MNL solution while the additional graph-dependent and nonlinear components are initialized at small magnitude. **Tables C.1** and **C.2** report the LOS coefficients, ASCs, and mode-specific individual-attribute coefficients for the SP and RP datasets. LOS-by-individual and pairwise interaction terms are included in the specification but not tabulated. The reference mode is bus on SP and walk on RP. Coefficients are reported without standard errors, since the interaction terms are L2-regularized and conventional unpenalized maximum-likelihood standard errors are not directly applicable.

**Table C.1.** MNL estimates for SP dataset (reference mode: bus).

| **Parameter** | **Estimate** | |
|---|---|---|
| **Level-of-service coefficients** | | |
| BUS_ACC | −0.139 | |
| BUS_WAIT | −0.160 | |
| BUS_TRANSFER | −0.303 | |
| SUB_ACC | −0.139 | |
| SUB_WAIT | −0.160 | |
| SUB_TRANSFER | −0.303 | |
| DRT_TT | −0.326 | |
| DRT_ACC | −0.139 | |
| DRT_WAIT | −0.160 | |
| DRT_TRANSFER | −0.303 | |
| DRT_FARE | −0.250 | |
| **Alternative-specific constants** | | |
| ASC_SUB | +0.342 | |
| ASC_DRT | +1.048 | |
| **Mode-specific individual-attribute coefficients, relative to bus** | | |
| | SUB | DRT |
| male | −0.03 | +0.01 |
| age_50 | +0.24 | +0.06 |
| edu_high | +0.07 | +0.20 |
| wage_300 | −0.06 | +0.10 |
| LICENSE | +0.08 | +0.09 |
| CAR | −0.13 | 0.00 |
| job_gen | +0.03 | +0.11 |
| job_pro | +0.18 | +0.03 |
| transfers_total | +0.13 | +0.16 |
| walk_15min | +0.18 | +0.15 |
| wait_15min | −0.27 | −0.04 |
| day_5 | +0.18 | +0.02 |
| comm_1 | +0.02 | −0.08 |
| F_punct | +0.08 | −0.17 |
| F_effic | −0.15 | +0.36 |
| F_comf | −0.64 | +0.05 |
| F_innov | 0.00 | +0.19 |

**Table C.2** MNL estimates for RP dataset (reference mode: walk).

| Parameter | | | Estimate |
|---|---|---|---|
| **Level-of-service coefficients** | | | |
| dur_walking | | | −1.276 |
| dur_cycling | | | −0.939 |
| dur_pt_access | | | −0.533 |
| dur_pt_rail | | | +0.031 |
| dur_pt_bus | | | −0.328 |
| dur_pt_int | | | −0.294 |
| pt_interchanges | | | −0.227 |
| pt_fare_per_km | | | −0.041 |
| dur_driving | | | −1.046 |
| cost_driving_ccharge | | | −0.363 |
| driving_traffic_percent | | | −0.437 |
| **Alternative-specific constants** | | | |
| ASC_cycle | | | −0.668 |
| ASC_PT | | | +2.009 |
| ASC_drive | | | +1.989 |
| **Mode-specific individual-attribute coefficients, relative to walk** | | | |
| | cycle | PT | drive |
| female | −0.44 | +0.15 | +0.07 |
| age | +0.02 | −0.15 | +0.07 |
| has_license | +0.12 | −0.35 | +0.28 |
| car_any | −0.18 | −0.19 | +1.01 |
| conc_age | −0.34 | +0.33 | +0.22 |
| conc_disab | +0.04 | −0.00 | −0.12 |
| distance_log | +1.58 | +2.82 | +2.45 |
| is_commute | +0.31 | +0.12 | −0.19 |
| rush_hour | +0.07 | +0.04 | +0.11 |
| weekend | −0.07 | −0.09 | +0.07 |
| winter | −0.21 | +0.00 | +0.01 |

The four public transport duration components in the RP specification are highly collinear, and the specification-matched MNL gives a small positive coefficient for rail duration (+0.031). Neural-BSL includes a sign penalty on the LOS coefficients and yields a negative mean coefficient for the same variable (−0.211). Rail duration is not directly changed in any of the scenarios in **Section 5.4**; the fare scenarios modify only the corresponding fare terms while rail duration remains at its observed value.

### C.2 MLP hyperparameter tuning

The MLP benchmark reported in **Section 5.1** is trained on the same input vector as the MNL and Neural-BSL specifications. Its architecture and regularization hyperparameters are selected by grid search over the number and width of hidden layers, the dropout rate, and the L2 weight decay at a fixed learning rate. Each of the twelve configurations is refit on the same five cross-validation folds used for the other models, and the one with the highest mean log-likelihood is reported. **Figure C.1** reports the mean log-likelihood across the configurations on each dataset.

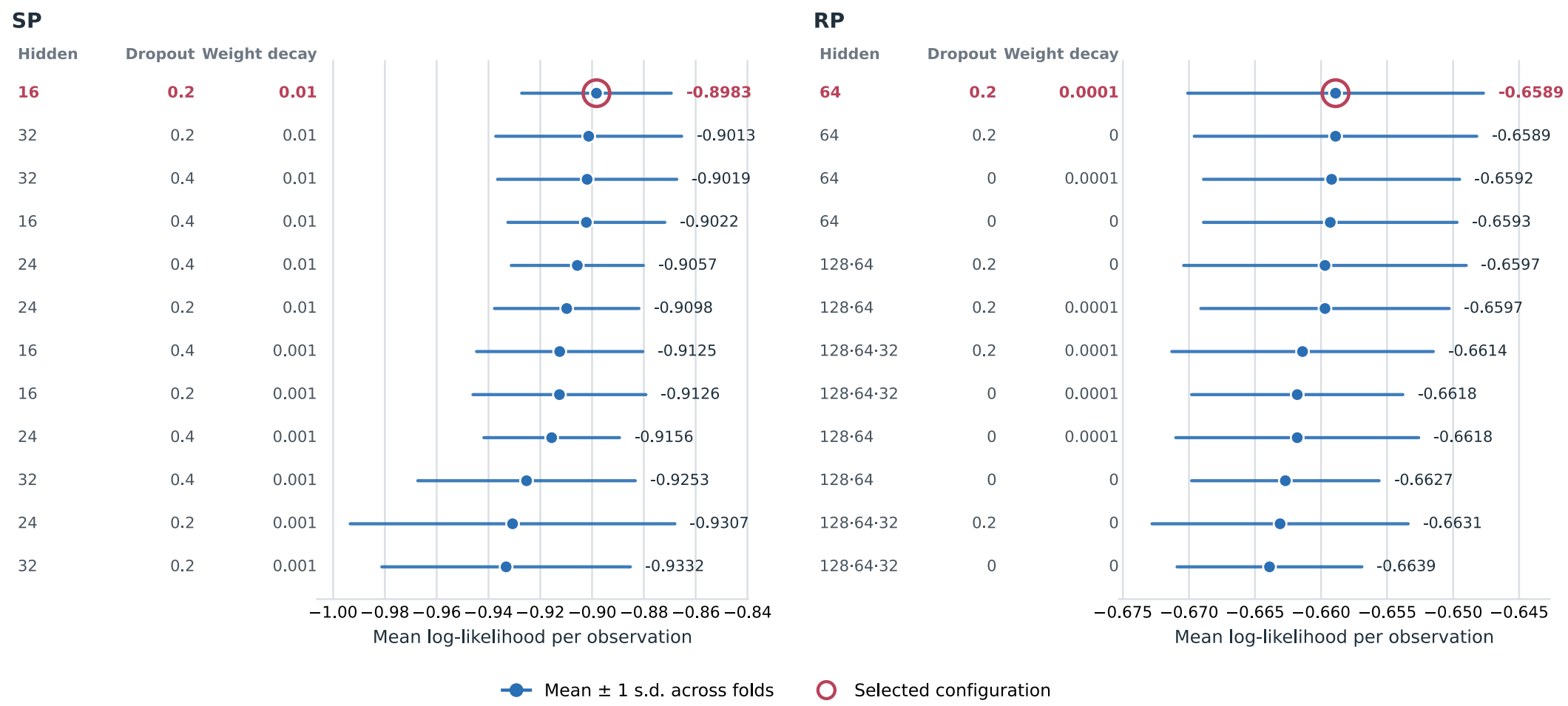

**Figure C.1.** MLP validation log-likelihood across the hyperparameter grid, SP (left) and RP (right).

On the SP data, the selected configuration is a single hidden layer of 16 units with dropout 0.2, learning rate 3e-3 and weight decay 1e-2 (mean log-likelihood −0.898). On the RP data, the selected configuration is a single hidden layer of 64 units with dropout 0.2, learning rate 1e-3, and weight decay 1e-4 (mean log-likelihood −0.659).

### C.3 Sensitivity analysis

We select the structural-loss weight $\lambda_{struct}$ by re-estimating Neural-BSL across a three-point grid on each dataset: $\lambda_{struct} \in \{1 \times 10^{-3}, 1 \times 10^{-2}, 5 \times 10^{-2}\}$ on SP and $\{1 \times 10^{-4}, 1 \times 10^{-3}, 1 \times 10^{-2}\}$ on RP. The grids differ because the structural and choice losses have different scales across the two datasets. **Figure C.2** reports test accuracy, held-out structural log-likelihood, and acyclicity violation across the candidate values. Held-out structural log-likelihood, averaged over the five folds used for hyperparameter optimization, improves as $\lambda_{struct}$ increases on both datasets. The acyclicity violation increases across the SP grid, while on RP it is smallest at $1 \times 10^{-4}$ and remains higher at the two larger values. Following the selection procedure in **Section 4.4**, we select $5 \times 10^{-2}$ on SP and $1 \times 10^{-4}$ on RP. Test accuracy changes only modestly across the grid, by approximately 0.4 percentage points on SP and 0.2 percentage points on RP, both within the cross-fold standard deviations reported in **Section 5.1**.

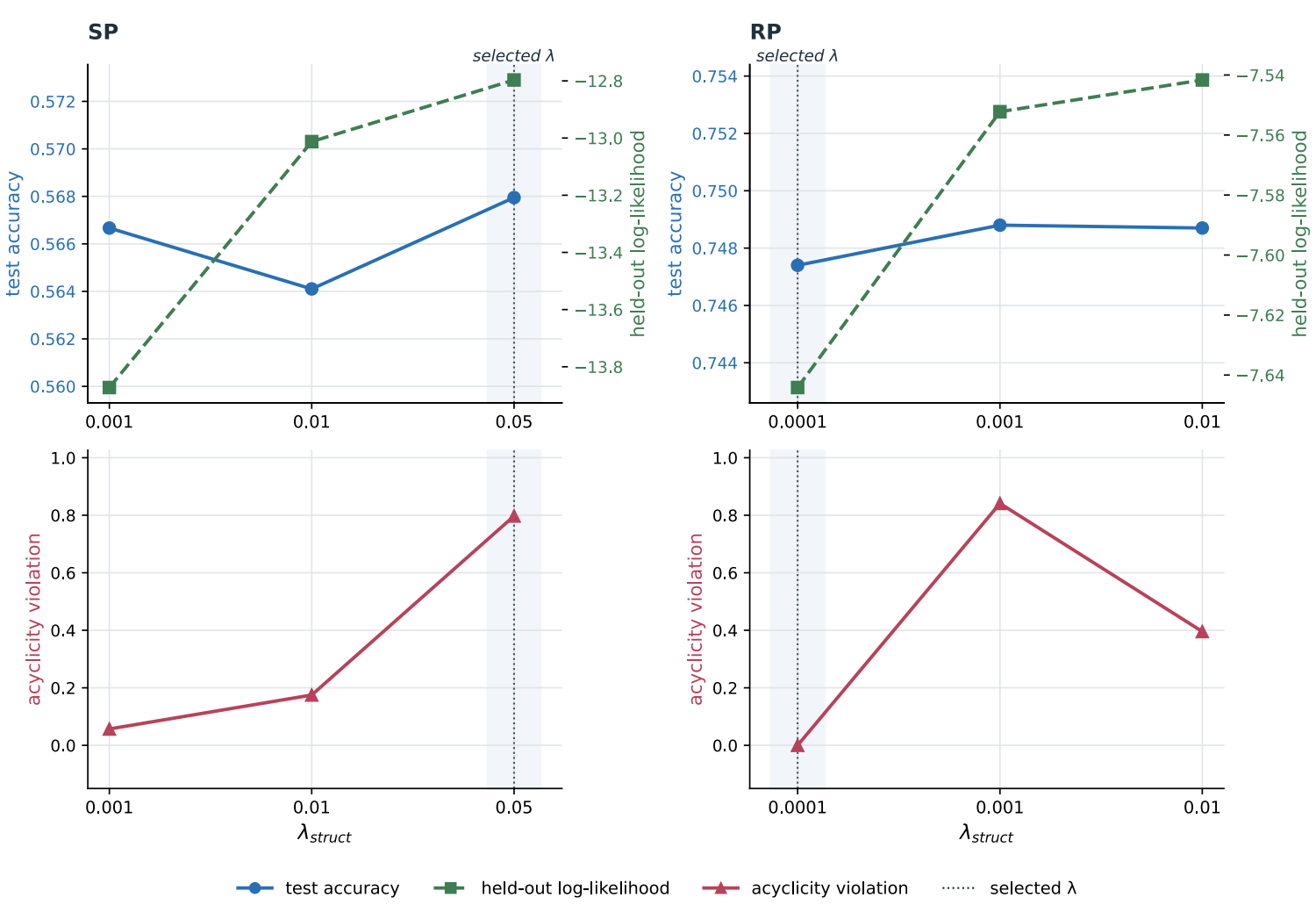

**Figure C.2.** Sensitivity of Neural-BSL to the structural-loss weight $\lambda_{struct}$, SP (left) and RP (right).